\documentclass[lettersize,journal]{IEEEtran}
\usepackage{amsmath,amsfonts}
\usepackage{algorithmic}
\usepackage{algorithm}
\usepackage{array}
\usepackage[caption=false,font=normalsize,labelfont=sf,textfont=sf]{subfig}
\usepackage{textcomp}
\usepackage{stfloats}
\usepackage{url}
\usepackage{verbatim}
\usepackage{graphicx}
\usepackage{cite}
\usepackage[colorlinks]{hyperref} 
\usepackage{graphicx}
\usepackage{amsmath}
\usepackage{ulem}
\usepackage{booktabs,amsfonts,dcolumn}
\usepackage{multirow}
\usepackage{wrapfig}
\usepackage{authblk}

\usepackage{algorithm}
\usepackage{listings}
\usepackage{color}

\usepackage{rotfloat}
\usepackage{lscape}
\usepackage[table,xcdraw]{xcolor}         

\begin{document}

\title{ProtoCLIP: Prototypical Contrastive \\ Language Image Pretraining}

\author{Delong Chen,
Zhao Wu, 
Fan Liu,~\IEEEmembership{Member,~IEEE,}
Zaiquan Yang,
Shaoqiu Zheng, 
Ying Tan,~\IEEEmembership{Senior Member,~IEEE,}
Erjin Zhou
\thanks{
This work was partially supported by National Nature Science Foundation of China (62372155), Joint Fund of Ministry of Education for Equipment Pre-research (8091B022123), Research Fund from Science and Technology on Underwater Vehicle Technology Laboratory (2021JCJQ-SYSJJ-LB06905), Key Laboratory of Information System Requirements, No: LHZZ 2021-M04, Water Science and Technology Project of Jiangsu Province under grant No. 2021063, Qinglan Project of Jiangsu Province.
Corresponding author: Fan Liu (e-mail: fanliu@hhu.edu.cn).}
\thanks{Delong Chen and Fan Liu are with the College of Computer and Information, Hohai University, Nanjing, China.}%
\thanks{Fan Liu is also with Science and Technology on Underwater Vehicle Technology Laboratory, Harbin, China.}%
\thanks{Delong Chen is also with Centre for Artificial Intelligence Research (CAiRE) of Hong Kong University of Science and Technology (HKUST), Hong Kong, China. This work is partially done when Delong Chen interned at MEGVII Research.}%
\thanks{Zhao Wu, Zaiquan Yang, and Erjin Zhou are with MEGVII Research, Beijing, China.}%
\thanks{Shaoqiu Zheng is with Nanjing Research Institute of Electronic Engineering.}%
\thanks{Ying Tan is with Department of Machine Intelligence, School of EECS, Peking University.}%
}


\maketitle

\begin{abstract}

Contrastive Language Image Pretraining (CLIP) has received widespread attention, since its learned representations can be transferred well to various downstream tasks. During the training process of the CLIP model, the InfoNCE objective aligns positive image-text pairs and separates negative ones. We show an underlying representation grouping effect during this process: the InfoNCE objective indirectly groups semantically similar representations together via randomly emerged within-modal anchors. Based on this understanding, in this paper, \textbf{Proto}typical \textbf{C}ontrastive \textbf{L}anguage \textbf{I}mage \textbf{P}retraining (ProtoCLIP) is introduced to enhance such grouping by boosting its efficiency and increasing its robustness against the modality gap. Specifically, ProtoCLIP sets up prototype-level discrimination between image and text spaces, which efficiently transfers higher-level structural knowledge. Further, \textbf{P}rototypical \textbf{B}ack \textbf{T}ranslation (PBT) is proposed to decouple representation grouping from representation alignment, resulting in effective learning of meaningful representations under large modality gap. The PBT also enables us to introduce additional external teachers with richer prior language knowledge. ProtoCLIP is trained with an online episodic training strategy, which makes it can be scaled up to unlimited amounts of data. We train our ProtoCLIP on Conceptual Captions and achieved an +5.81\% ImageNet linear probing improvement and an +2.01\% ImageNet zero-shot classification improvement. {On the larger YFCC-15M dataset, ProtoCLIP matches the performance of CLIP with 33\% of training time.}
\end{abstract}


\section{Introduction}\label{sec:introduction}

    \IEEEPARstart{C}{ontrastive} Language Image Pretraining (CLIP)~\cite{Radford2021Learning} has achieved impressive performance on learning representations from large-scale image-text pairs collected from the Internet. 
    It optimizes the Information Noise Contrastive Estimation (InfoNCE) objective~\cite{Oord2018Representation} during pretraining, but how this simple objective derives meaningful image-text representations is not well studied. Intuitively, the InfoNCE objective creates a joint representation space, where paired image-text representations are pushed together and unpaired representations are pulled apart. This property can be termed as \textit{representation alignment}. {However, fulfilling this alone is inadequate for perfect downstream performance. For example, the image-text representations can be perfectly aligned \cite{Wang2020Understanding} and randomly distributed at the same time. In such a situation, the InfoNCE objective can still reach its minimum, but downstream performance would be poor \cite{Wang2022Chaos}.}
    
    \begin{figure*}
      \centering
      \includegraphics[width=0.95\linewidth]{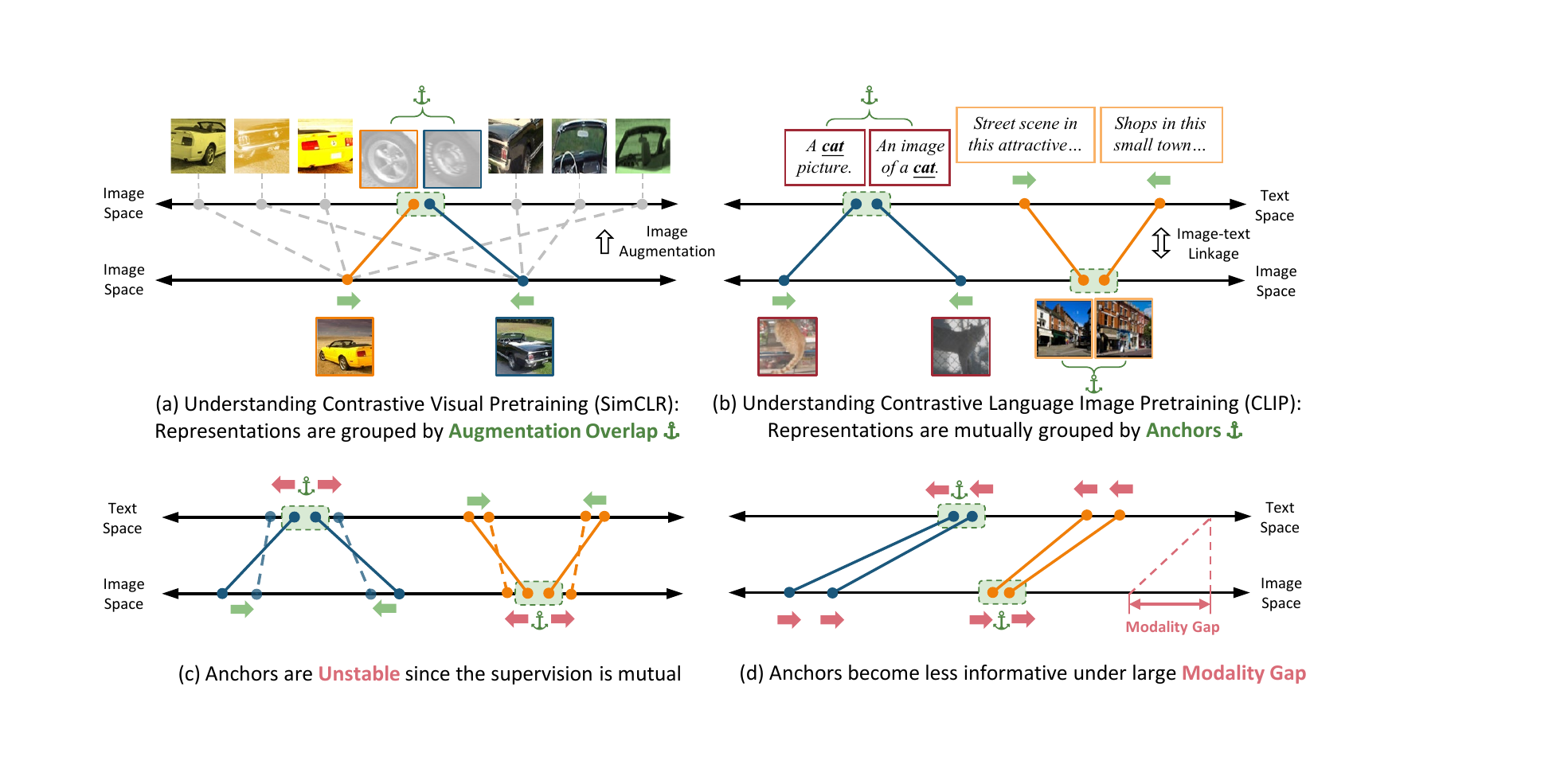}
      \caption{Illustrations of representation grouping in 1-dimensional spaces. Each ``$\bullet$---$\bullet$'' represents an image-image (a) or image-text representation (b)-(d) pair.}
      \label{fig:2d_feature_space}
    \end{figure*}
    

    This contradiction motivated us to seek a new understanding of CLIP's learning process beyond representation alignment. {We find the recent ``augmentation overlap''~\cite{Wang2022Chaos} theory is particularly instructive for this. As shown in Figure~\ref{fig:2d_feature_space}(a), in contrastive visual pretraining (e.g., SimCLR~\cite{Chen2020Simple}), aggressive image augmentations are randomly applied to generate different views. The ``augmentation overlap'' \cite{Wang2022Chaos} theory suggests that pairs of indistinguishable views (e.g, the wheels of two different cars, noted by the green dashed box) will emerge during this process. When the InfoNCE objective aligns different views together, these overlapped augmentations will group intra-class samples together (as noted by green arrows).}
    {By regarding different modalities as different views, we can naturally extend the above theory to a multimodal setting. As in Figure~\ref{fig:2d_feature_space}(b), now the inherent linkage of image-text pairs plays a similar role to image augmentations.} Close within-modality pairs (noted by green dashed boxes) will emerge and serve as ``anchors'' to group the corresponding representations in the opposite modality. For example, robust visual features (e.g., recognizing cats from different angles) can be learned through the cooccurrence of the word ``cat'' in a pair of text captions.
    Such \textit{representation grouping} of InfoNCE has been demonstrated to be effective, but our new ``anchor-grouping'' understanding reveals its two main weaknesses.
    \textbf{First}, {the grouping is done in an \textit{indirect} manner, anchors are prone to be pulled apart by the ``reaction'' of the immature opposite modality. When the text anchor in Figure~\ref{fig:2d_feature_space}(b) pushes two cat images together (i.e., learning robust visual representations as desired, as green arrows in Figure~\ref{fig:2d_feature_space}(c)), the large distance between these immature image representations would separate the text anchor apart (for example, learning to discriminate word ``picture'' and ``image'' undesirably, as red arrows in Figure~\ref{fig:2d_feature_space}(c)).} Such ``reaction'' leads to a reduced number of effective anchors and yields less grouped representations.
    \textbf{Second}, anchors become less informative with the existence of a large modality gap. Modality gap~\cite{Liang2022Mind,Bai2022LaT} is defined as the range between the mean representations in the image and text spaces. As shown in Figure~\ref{fig:2d_feature_space}(d), when the two representation spaces are not overall aligned, the InfoNCE objective will focus primarily on aligning them to minimize the modality gap rather than learning meaningful representations via anchor-grouping {since the large gap overwhelms the relational information within each modalities. A similar problem has been well explored in the field of knowledge distillation~\cite{Hinton2015Distilling}, where researchers found that the ``absolute teacher'' is not robust to representation space translation---a single modal version of modality gap---and yields sub-optimal performance~\cite{Yu2019Learning}. Unfortunately, at the beginning of CLIP training, a large modality gap occurs with a very high possibility due to the independent initialization of CLIP's two encoders and the inherent ``cone effect'' of non-linear deep neural networks~\cite{Liang2022Mind}. } 
    
\begin{figure}
      \centering
      \includegraphics[width=\linewidth]{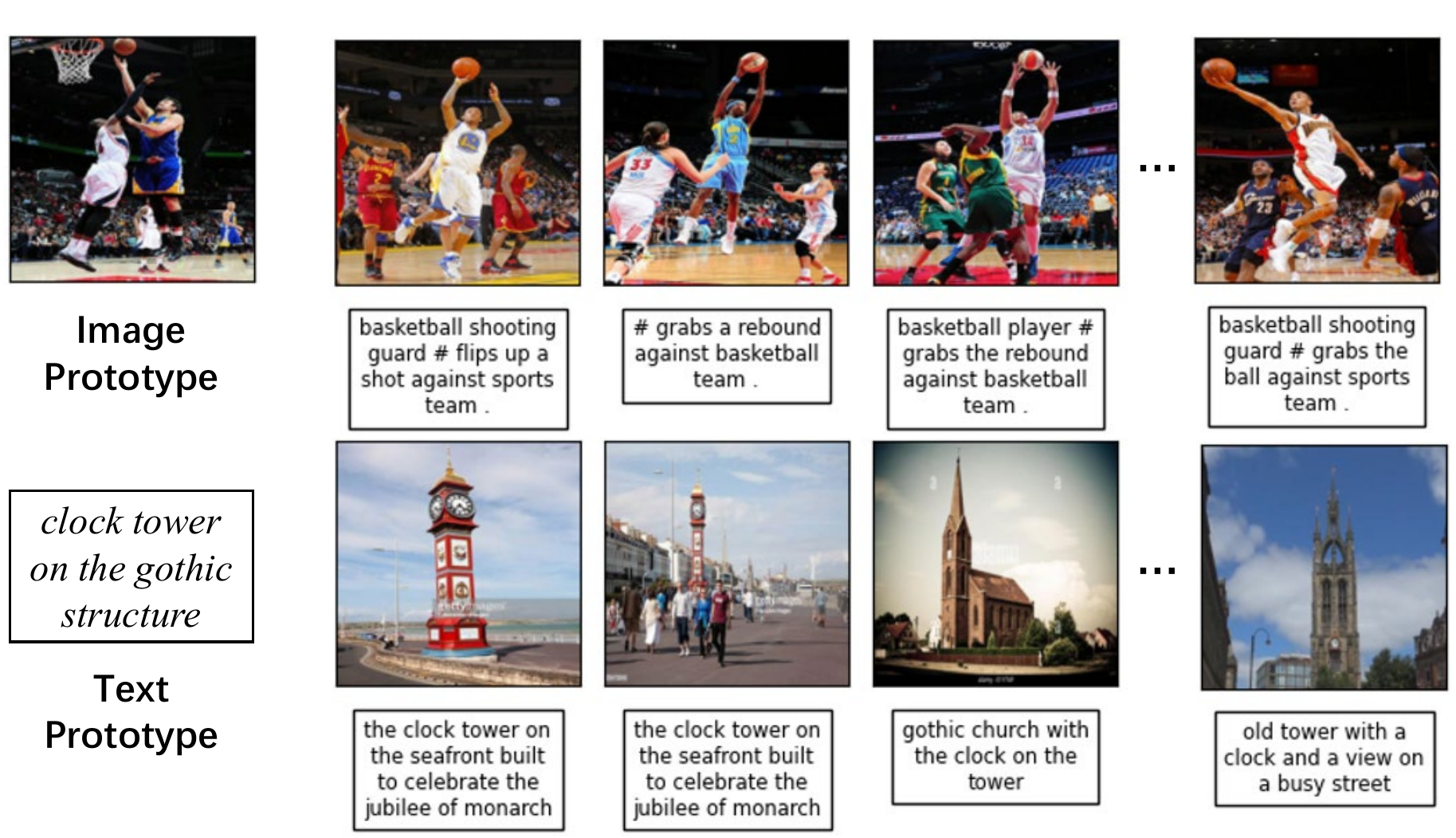}
      \caption{{\textbf{Left}: Image and text prototypes recognized by ProtoCLIP. Each prototype represents a high-level semantic units. \textbf{Right}: samples assigned to the corresponding prototype, they have similar semantics with the prototypes.}}
      \label{fig:prototype_visuallization}
    \end{figure}

    We propose \textbf{Proto}typical \textbf{C}ontrastive \textbf{L}anguage \textbf{I}mage \textbf{P}retraining (ProtoCLIP), which raises instance-level discrimination to prototype-level discrimination by constructing and dynamically updating prototypes on both image-text spaces. {As shown in Figure~\ref{fig:prototype_visuallization}, samples assigned to the same prototype have shared semantics, and we use these prototypes to \textit{directly} supervise the opposite modality. This leads to richer supervision signals and more efficient representation grouping. Prototypical supervisions are comparatively more stable, since these prototypes do not have the risk of being pulled apart.}
    
    For the modality gap, we further introduce a simple yet effective \textbf{P}rototype \textbf{B}ack \textbf{T}ranslation (PBT) technique to decouple representation grouping from representation alignment. PBT calculates a within-modal centroid for samples that are assigned to a shared prototype, and then groups these representations towards the centroid. With PBT, representation alignment is no longer a prerequisite for effective learning of representation grouping. Based on the ability to learn representations from unaligned spaces, we can further introduce an external teacher (e.g., a pre-trained RoBERTa~\cite{Liu2019RoBERTa}) with richer prior knowledge.
    
    Furthermore, we present two improvements to previous clustering-based pretraining methods. 
    \textbf{First}, DeepCluster~\cite{Caron2018Deep, Caron2020Unsupervised}, SeLa~\cite{Asano2020Self}, PCL~\cite{Li2021Prototypical}, XDC~\cite{Alwassel2020Self}  SeLaVi~\cite{Asano2020Labelling}, and MCN~\cite{Chen2021Multimodal} 
    update the clusters after each training epoch or several consecutive epochs. Such a training strategy can work well on medium-sized ImageNet~\cite{Deng2009ImageNet} but is not scalable to larger datasets (e.g., YFCC~\cite{Thomee2016YFCC100M}) due to low cluster updating frequency. To train ProtoCLIP more efficiently, we design an online episodic training strategy, which makes the training of ProtoCLIP can be scaled up to unlimited amounts of data. \textbf{Second}, previous works~\cite{Caron2018Deep, Asano2020Self, Alwassel2020Self, Asano2020Labelling, Li2021Prototypical, Zhang2021Supporting} learn one-hot pseudo labels as hard targets, which ignores the structural relationship among clusters. 
    For example, although the ``cat'' and ``tiger'' samples probably belong to different clusters, the distance between them should be much closer than that of ``cat'' and ``car''.
    To this end, we convert hard cluster assignments to probability scores using softmax to enable the effective transfer of such relational knowledge.
    Overall, our main contributions in this paper are summarized as follows:
    
    \begin{itemize}
        \item We proposed ProtoCLIP with prototype-level discrimination that enables more efficient representation grouping in large-scale vision-language pretraining. {Prototypes serve as stable anchors to group the representations of semantically similar samples together.}
        
        \item We designed PBT to translate cross-modal prototypes to within-modal centroids. PBT enables ProtoCLIP to learn meaningful representations between unaligned spaces. Via PBT, we further introduced pretrained RoBERTa as an external teacher for richer supervision.
        
        \item We presented two improvements to previous clustering-based pretraining methods: 1) an online episodic training strategy that improves cluster updating frequency, and 2) the use of probability-based soft targets which transfer structural relational knowledge.
        
        \item Experimental results in Conceptual Captions 3M showed that ProtoCLIP outperforms CLIP by +5.81\% and +2.01\% on ImagNet linear probing and zero-shot classification respectively. {On the larger YFCC dataset, ProtoCLIP matched the performance of CLIP with 33\% train time cost.} Codes are available at \url{https://github.com/megvii-research/protoclip}.
    \end{itemize}

\section{Related Works}
\subsection{Vision Language Pretraining.} 
    {Recent works have exploited learning multimodal representations from large-scale web-crawled image-text data and showed promising results. 
    The amount of training data can be scaled up to even hundreds of millions of samples or even billions of samples, which poses strong regularization that prevents overfitting and enables the model to learn open-vocabulary visual concepts. VLP models can be classified into single-stream and dual-stream:
    
    \textbf{Single-stream models}~\cite{Li2019VisualBERT, Lu2019ViLBERT, Chen2020UNITER, Li2020Oscar,Chen2023Visual} fuse image and text based on the advantage of the attention mechanism~\cite{Vaswani2017Attention} and excel at multimodal fusion and understanding, leading to impressive performance in multimodal high-level tasks such as Visual Question Answering (VQA) and image captioning. Unfortunately, the transferability of single-stream models is weak, since they have no independent encoder that can be transferred to single-modal tasks.
    
    \textbf{Dual-stream models} set up two separate encoders to align visual and textual representations. Although the methodology is quite simple, pioneering work~\cite{Radford2021Learning}) demonstrated prestigious success when combining it with a huge amount of training data and large vision Transformers~\cite{Liu2023Survey}. Some follow-up work improved CLIP from the representation alignment perspective. For example, FILIP~\cite{Yao2021FILIP} introduced f0iner-grained representation alignment to boost multimodal interaction. CLOOB~\cite{Fuerst2021CLOOB} introduced Hopfield Networks for improved learning of feature associations and cooccurrences. More recent efforts have focused on improving learning efficiency, since CLIP training is highly expensive. To improve learning efficiency, EfficientCLIP~\cite{Wang2021EfficientCLIP} and SLIP~\cite{Mu2021SLIP} respectively combined BERT~\cite{Devlin2019BERT}-style and SimCLR~\cite{Chen2020Simple}-style single-modal self supervision with CLIP. DeCLIP~\cite{Li2021Supervision} further integrates multi-view supervision and nearest-neighbor supervision. RemoteCLIP~\cite{Liu2023RemoteCLIP} adapts CLIP to remote sensing domain.

\subsection{Self-supervised Visual Representation Learning.} 
    Self-supervised Learning (SSL)~\cite{Ericsson2021Self} aims at learning meaningful representations without human supervision. Early works on SSL focus on exploring pretext tasks~\cite{Jing2021Self}. After SimCLR~\cite{Chen2020Simple} demonstrated the effectiveness of instance discrimination task, contrastive learning became dominant. SimCLR aligns representations of different data augmentations, which creates augmentation overlaps~\cite{Wang2022Chaos} that groups intra-class samples together. Unfortunately, SimCLR relies on extremely large batch sizes for sufficient negatives. To solve this issue, MoCo~\cite{He2020Momentum} introduced momentum contrast, while BYOL~\cite{Grill2020Bootstrap} and SimSiam~\cite{Chen2021Exploring} showed that representations can be learned without negatives. Though these works effectively improved SSL learned representations, they share a fundamental weakness that the model is only encouraged to learn augmentation-invariant representations, while higher levels of semantic relations are ignored. Nearest Neighbor-based methods such as NNCLR~\cite{Dwibedi2021Little} and MYOL~\cite{Azabou2021Mine} introduced richer supervision signals, but the variance of positive pairs is still limited. 

\subsection{Clustering-based SSL.} 
    A promising line of work in SSL is clustering-based approaches. DeepCluster~\cite{Caron2018Deep} and SeLa~\cite{Asano2020Self} assign pseudo labels using \textit{K}-Means or Sinkhorn Knopp algorithm, then use these labels to supervise model training. SwAv~\cite{Caron2020Unsupervised} contrasts the cluster assignment between different augmentations of the same image. The clustering of SwAV is done in an online fashion, but it forces the size of each cluster to be equal. PCL~\cite{Li2021Prototypical} and SCCL~\cite{Zhang2021Supporting} combined cluster-level contrast with instance-level contrast and demonstrated the effectiveness in image SSL and text SSL respectively. In Prototypical Graph Contrastive Learning (PGCL)~\cite{Lin2022Prototypical}, prototype-level contrastive learning is introduced for SSL on graph data. In~\cite{Wang2022Meta}, meta-prototype learning is proposed to improve few-shot image recognition.
    
    Clustering-based learning on multi-modal data is an emerging topic. XDC~\cite{Alwassel2020Self} and SeLaVi~\cite{Asano2020Labelling} respectively extend DeepCluster~\cite{Caron2018Deep} and SeLa to audio-visual pretraining~\cite{Asano2020Self}. {XDC~\cite{Alwassel2020Self} also conducted an extensive comparison of different types of supervision (i.e., single-modal v.s. multi-modal fusion v.s. cross-modal). They found that all of them are effective and a model learns the best when it is purely supervised by the opposite modality. Inspired by XDC, ProtoCLIP also creates cross-modal supervision in a cross-modal manner. We empirically found that multi-modal fusion-based supervision (i.e., the CDC~\cite{Alwassel2020Self}) yields significantly degenerated performances for VLP. The density of initial random text representations is much higher than that of image representations, which makes it dominate the pseudo label generation and fails to learn useful knowledge from the image representations.}
    
    ProtoCLIP shares some similarities with XDC~\cite{Alwassel2020Self}, since both of them utilize the clusters in the opposite modality as supervision. However, ProtoCLIP aims at VLP instead of audio-visual pretraining which only requires representation grouping---in a VLP scenario, representation alignment should be considered as well for zero-shot classification and cross-modal retrieval. Additionally, compared to a pure VLP version of XDC, ProtoCLIP contains several novel designs, including PBT, episodic training, learnable temperature, and the use of soft targets.

\section{Method}

\subsection{Prototypical Contrastive Language Image Pretraining}

    Let's get started by revisiting the InfoNCE objective used by the original CLIP~\cite{Radford2021Learning}. CLIP is trained with a large-scale image-text dataset $\mathcal{D}=\{(x^I_i, x^T_i)\}_{i=1}^M$ that consists of a total of $M$ training samples. The goal is to learn an image encoder $f^I$ and a text encoder $f^T$ that respectively encode $x^I_i$ and $x^T_i$ to their latent representations, i.e., $f^I(x^I_i) = z^I_i\in\mathbb{R}^{d_z \times 1}$ and  $f^T(x^T_i) = z^T_i\in\mathbb{R}^{d_z \times 1}$. The learned representation should fulfill two requirements: representation alignment and representation grouping:
    
    \textbf{Representation alignment} refers to high similarity $z^I_i \cdot z^T_i$  of paired image and text samples $x^I_i, x^T_i$, and low similarity $z^I_i \cdot z^T_j (i \neq j)$ between the unpaired samples $x^I_i, x^T_j$. Generally, perfect representation alignment yields strong downstream performance on cross-modal retrieval tasks.
        
    \textbf{Representation grouping} means that representations of semantically similar samples are grouped together, while those of dissimilar samples should be pulled apart. Perfect representation grouping yields strong linear classification performance. 
    
    While fulfilling perfect representation alignment and representation grouping at the same time, coupled with a large dataset that contains sufficient open-set concepts, the model can achieve strong zero-shot classification performance. To achieve this objective, CLIP creates an instance discrimination task within each batch, and optimizes the following bidirectional InfoNCE objective~\cite{Oord2018Representation} to maximize the mutual information between paired image and text~\cite{Tu2023Hierarchically}.

    \begin{equation}\label{eq:loss_CLIP}
    \begin{aligned}
        \mathcal{L}_{\text{CLIP}}
        = &- (
        \underbrace{
        \frac{1}{N} \sum_{i=1}^{N} \log \frac{\text{exp}(z^I_i \cdot z^T_i/\tau_{\text{CLIP}})}{\sum_{j=1}^{N}\text{exp}(z^I_i \cdot z^T_j/\tau_{\text{CLIP}})}}_{\text{image to text}} \\
        &+ \underbrace{\frac{1}{N} \sum_{i=1}^{N} \log \frac{\text{exp}(z^T_i \cdot z^I_i/\tau_{\text{CLIP}})}{\sum_{j=1}^{N}\text{exp}(z^T_i \cdot z^I_j/\tau_{\text{CLIP}})}}_{\text{text to image}})/2,
    \end{aligned}
    \end{equation}

Here, $N$ is the batch size and $\tau_{\text{CLIP}}$ is the temperature parameter. Following the CLIP paper, we set it as a learnable parameter.
    As illustrated in Section~\ref{sec:introduction}, representation grouping is done \textit{indirectly} by the InfoNCE objective. Here, we want to boost the efficiency by performing representation grouping in a \textit{direct} manner. We raise the instance-level discrimination to prototype-level discrimination by constructing and updating prototypes. A prototype is a representation for a group of semantically similar instances~\cite{Li2021Prototypical}. We push representations of the prototype directly for grouping through a proposed prototypical loss $\mathcal{L}_{\text{Proto}}$. 
    
    {An illustration of ProtoCLIP architecture is shown in Figure \ref{fig:model}. To acquire prototypes, we apply MLP projection heads $g^I$ and $g^T$ on top of $z^I_i$ and $z^T_i$ respectively, then we get projected representations $g^I(z^I_i)=h^I_i\in\mathbb{R}^{d_h \times 1}$ and $g^T(z^T_i)=h^T_i\in\mathbb{R}^{d_h \times 1}$. Prototypes are constructed here in the projected representation spaces ($h^I_i$ and $h^T_i$) instead of the raw representation spaces ($z^I_i$ and $z^T_i$). This is done for two reasons. First, we want ProtoCLIP to hold the instance-level discrimination ability of CLIP by keeping the $\mathcal{L}_{\text{CLIP}}$, so prototypical-level discrimination should be done elsewhere otherwise it will cause conflicts between $\mathcal{L}_{\text{CLIP}}$ and $\mathcal{L}_{\text{Proto}}$. Second, the MLP projection heads $g^I$ and $g^T$ can project representations to lower-dimensional spaces (i.e. $d_h<d_z$), such that the cost of constructing prototypes can be significantly reduced.}

    We adopt \textit{K}-Means clustering due to its simplicity and scalability. Other clustering methods can be used here as well. Specifically, we find prototypes $C^I\in\mathbb{R}^{K \times d_h} = [c^I_1, c^I_2, ..., c^I_K]$ and $C^T\in\mathbb{R}^{K \times d_h} = [c^T_1, c^T_2, ..., c^T_K]$ that minimize the following \textit{K}-Means objective:

    \begin{figure*}
      \centering
      \includegraphics[width=0.7\linewidth]{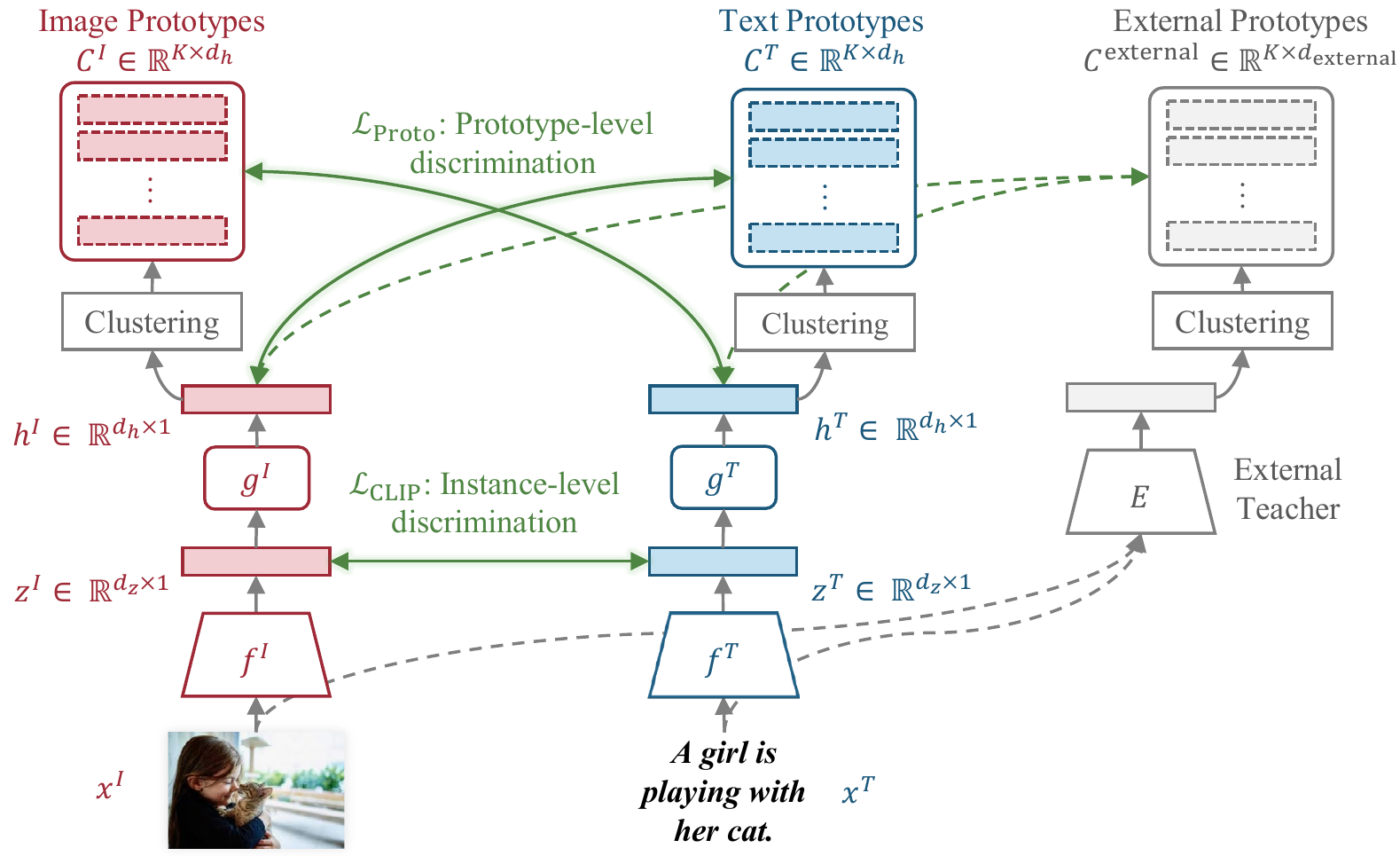}
      \caption{\textbf{Model Architecture of ProtoCLIP.} We setup prototype-level discrimination upon the instance-level discrimination. We construct prototypes with representations after projection heads $g^I$, $g^T$. The prototypes are used to guide the learning of the opposite modality. An external teacher $E$ is introduced for richer supervision, which will be detailed in Section~\ref{sec:learning_unaligned}.}
      \label{fig:model}
    \end{figure*}

    \begin{equation}
    \arg \min_{C^I, C^T} 
    \underbrace{\sum^{K}_{k=1} \sum^{M}_{i=1} \left \| g^T(z^T_i) - c^T_k \right \|^2}_{\text{clustering text representations}} + 
    \underbrace{\sum^{K}_{k=1} \sum^{M}_{i=1} \left \| g^I(z^I_i) - c^I_k \right \|^2}_{\text{clustering image representations}}
    \end{equation}
    
    {Note that this process is operated on detached representations, so that the network would not receive any gradient.} After \textit{K}-Means clustering, pseudo labels (or cluster assignment) can be then generated for each sample according to the closeness between its representation and each prototype. 
    Previous clustering-based audio-visual pretraining method XDC~\cite{Alwassel2020Self} have compared different types of supervision and found model learns the best when it is purely supervised by the opposite modality. Inspired by XDC, Here ProtoCLIP creates prototypical supervision in a cross-modal manner: we use the prototypes in the opposite modality to guide representation learning. We empirically found that multi-modal fusion-based supervision (i.e., the CDC~\cite{Alwassel2020Self}) yields significantly degenerated performances for VLP. The density of initial random text representations is much higher than that of image representations, which makes it dominate the pseudo label generation and failed to learn useful knowledge from the image representations.
    Besides, previous methods such as DeepCluster~\cite{Caron2018Deep} and XDC~\cite{Alwassel2020Self} generate class indices and train an additional parametric classifier with cross-entropy loss, as usually done in traditional supervised training. However, since there is no mapping between two consecutive cluster assignments, such a method requires frequent re-initialization of the classifier, which interrupts the training procedure. Instead, we use the prototypes as linear classifiers directly~\cite{Caron2020Unsupervised, Li2021Prototypical}. As in Eq. \ref{eq:student_score}, we calculate classification scores $S^T_i\in\mathbb{R}^{k \times 1}$ and $S^I_i\in\mathbb{R}^{k \times 1}$  by applying the prototype classifier to the cross-modal representations, then normalize the scores to possibilities by taking softmax:

    \begin{equation}\label{eq:student_score}
    \begin{aligned}
       & p^I_i = \text{softmax}(S^I_i / \tau_{\text{Proto}}),& \quad\quad\text{where}\quad &S^I_i = C^T \cdot h^I_i. \\
       & p^T_i = \text{softmax}(S^T_i / \tau_{\text{Proto}}),& \quad\quad\text{where}\quad &S^T_i = C^I \cdot h^T_i.
    \end{aligned}
    \end{equation}
    
    where $\tau_{\text{Proto}}$ is the temperature hyper-parameter. Instead setting a fixed temperature as in DeepCluster-v2~\cite{Caron2020Unsupervised}, we set it to a learnable parameter as in $\mathcal{L}_{\text{CLIP}}$ since it yields improved results. Now, we can get $\mathcal{L}_{\text{Proto}}$ by applying the cross entropy loss function:

    \begin{equation}\label{eq:loss_Proto}
        \mathcal{L}_{\text{Proto}}  
        = -(
        \underbrace{\sum_{i=1}^{M}\sum_{k=1}^{K} y^T_{i,k} \log (p^I_{i,k})}_{\text{image to text}}
        + \underbrace{\sum_{i=1}^{M}\sum_{k=1}^{K} y^I_{i,k} \log (p^T_{i,k})}_{\text{text to image}})/2,
    \end{equation}
    
    \subsection{Learning from Soft Targets}
    In Eq.~\ref{eq:loss_Proto}, $y^T_i\in\mathbb{R}^{k \times 1}$ and $y^I_i\in\mathbb{R}^{k \times 1}$ are k-way pseudo target scores. Previous clustering-based methods~\cite{Caron2018Deep, Asano2020Self, Li2021Prototypical, Alwassel2020Self, Asano2020Labelling} convert class indices to a one-hot vector as targets. Such a one-hot target creates a one-vs-all learning task: representations are pushed towards their assigned prototypes \textit{only} and pushed away from other prototypes \textit{equally}.    
    To learn more structured knowledge, we use probability-based soft targets to replace the hard one-hot assignment:
    
    \begin{equation}\label{eq:teacher_score}
    \begin{aligned}
        y^T_k = \text{softmax}(S^{\hat{T}}_k / \tau_{\text{y}}),&\quad\quad\text{where}\quad S^{\hat{T}}_k = C^T \cdot c^T_k. \\
        y^I_k = \text{softmax}(S^{\hat{I}}_k / \tau_{\text{y}}),&\quad\quad\text{where}\quad S^{\hat{I}}_k ~=~ C^I \cdot c^I_k.
    \end{aligned}
    \end{equation}
    
    The scores in Eq.~\ref{eq:teacher_score} are calculated by measuring the dot-product similarity between the ``ground truth'' prototype $c^I_k, c^T_k$ and all the prototypes $C^I, C^T$. The ``ground truth'' prototype will have the highest similarity with itself (e.g., ``cat'' and ``cat''), relatively high similarities with its neighboring prototypes (e.g., ``cat'' and ``tiger''), and low similarities with distant prototypes (e.g., ``cat'' and ``car''). When such relational knowledge is embedded in the targets $y^I_k, y^T_k$, the ProtoCLIP can learn more structured knowledge. 
    Finally, the ProtoCLIP is trained to minimize $\mathcal{L}_{\text{Proto}}$ and $\mathcal{L}_{\text{CLIP}}$ jointly:
    
    \begin{equation}
        \mathcal{L}_{\text{ProtoCLIP}} =  \mathcal{L}_{\text{Proto}} + \mathcal{L}_{\text{CLIP}}
    \end{equation}

\subsection{Learning Representation Grouping from Unaligned Spaces}\label{sec:learning_unaligned}
    
    We compare the differences between $\mathcal{L}_{\text{CLIP}}$ and $\mathcal{L}_{\text{Proto}}$ in Figure~\ref{fig:teacher_student_spaces}(a) and (b). Though $\mathcal{L}_{\text{Proto}}$ improves the representation grouping efficiency, it still suffers from the \textit{modality gap} problem. In Figure~\ref{fig:teacher_student_spaces} (b), all the three data points in the student space would be pushed to the right side in order to align them with the prototype in teacher space.
    
    \begin{figure*}[!t]
      \centering
      \includegraphics[width=0.8\linewidth]{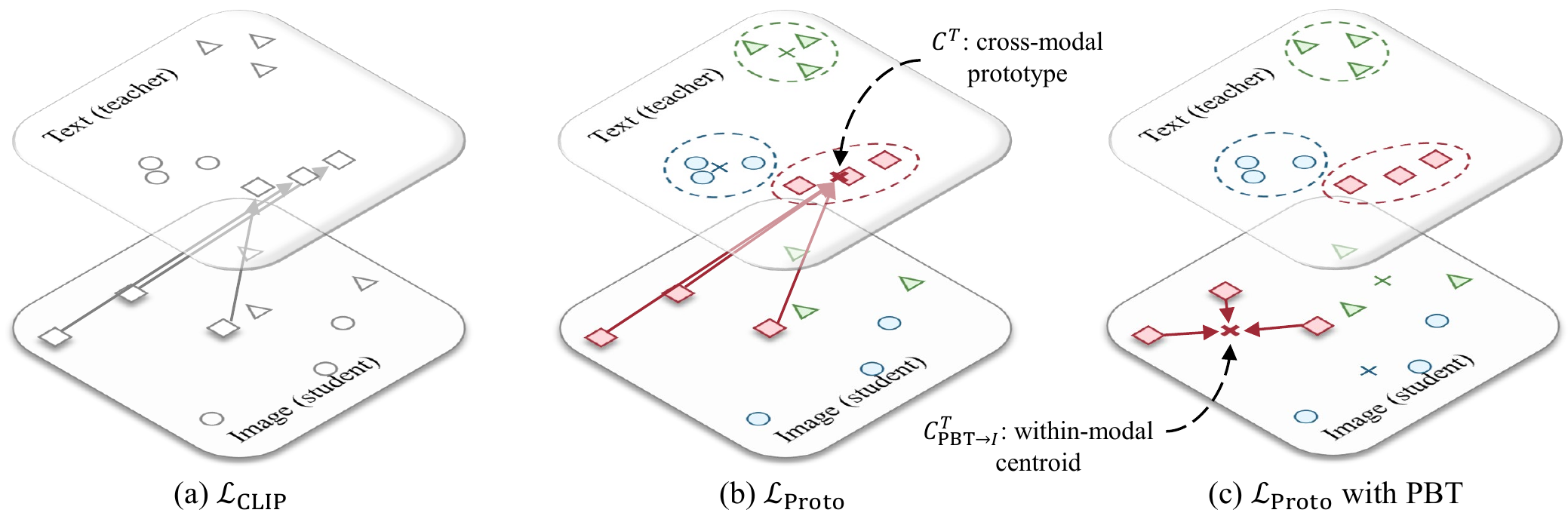}
      \caption{\textbf{Comparison of $\mathcal{L}_{\text{CLIP}}$, $\mathcal{L}_{\text{Proto}}$, and  $\mathcal{L}_{\text{Proto}}$ with PBT}. Our PBT translates cross-modal prototypes ($C^T$) to within-modal centroids ($C^T_{\text{PBT} \to I}$) according to prototype assignment. Since both of these losses are bi-directional between image and text spaces, here we only visualize the supervision from text (as teacher) to image (as student).}
      \label{fig:teacher_student_spaces}
    \end{figure*}

\subsubsection{Prototype Back Translation (PBT)}

The core reason of the modality gap problem is that $\mathcal{L}_{\text{Proto}}$ forces student representations to be strictly anchored to the position of their prototype in the teacher space. We introduce a simple yet effective technique called Prototype Back Translation (PBT) to avoid this problem. As shown in Figure~\ref{fig:teacher_student_spaces}(c), for each prototype in teacher space, we retrieve all the samples that are assigned to it, and then calculate a centroid of the corresponding representations in the student space. We denote the obtained image and text centroids as $C^T_{\text{PBT} \to I}$ and $C^I_{\text{PBT} \to T}$ and use them to replace the original prototypes $C^T$ and $C^I$ when calculating $\mathcal{L}_{\text{Proto}}$. Specifically, the computation of $C^T_{\text{PBT} \to I}$ involves the following two steps: 

\begin{enumerate}
    \item Retrieving assigned image samples $\mathcal{I}_k$ for each $c^T_k$ in $C^T$: For each prototype $c^T_k$ in $C^T$, identify the set of image samples $\mathcal{I}_k$ that are assigned to this prototype.
    \item Calculating centroid $c^T_{\text{PBT} \to I, k}$ of $\mathcal{I}_k$ representations: Calculate the centroid $c^T_{\text{PBT} \to I, k}$ of the representations of the assigned image samples $\mathcal{I}_k$ in the student image space. The centroid is computed as the average of the individual representations:
\end{enumerate}

   \begin{equation}
c^T_{\text{PBT} \to I, k} = \frac{1}{|\mathcal{I}_k|} \sum_{i \in \mathcal{I}_k} h^I_i
   \end{equation}

Here, $h^I_i$ represents the projected representation of the image sample $i$. This centroid $c^T_{\text{PBT} \to I, k}$ serves as a replacement for the original prototype $c^T_k$ in the computation of the prototypical loss $\mathcal{L}_{\text{Proto}}$. Similar steps are followed for computing $C^I_{\text{PBT} \to T}$ in the text space.

PBT enables knowledge transfer between unaligned representation spaces, since student representations are grouped directly to their within-modal centroid instead of pushed towards their cross-modal prototypes. We note that the advantages of $\mathcal{L}_{\text{Proto}}$ + PBT over plain $\mathcal{L}_{\text{Proto}}$ are similar to the advantages of Relational Knowledge Distillation (RKD)~\cite{Park2019Relational, Yu2019Learning} over traditional Knowledge Distillation (KD)~\cite{Hinton2015Distilling}. However, RKD transfers relational knowledge in a per-sample pair manner, while PBT transfers knowledge via prototypes with a higher level of semantics. Furthermore, PBT shares some similarity with the RACG model for imgae text retrieval~\cite{Peng2022Relation}, which also explicitly learns fine-grained semantic correspondence from intermodal representation relationships.

\subsubsection{Learning from External Teacher} 
    Since representation grouping is decoupled from representation alignment, we can now ensemble multiple teachers to guide the learning of student representations. For example, in addition to the original mutual knowledge transfer between image and text spaces, we can further introduce an external teacher encoder $E$ to distill richer knowledge to ProtoCLIP. As Figure~\ref{fig:model}, the encoder $E$ can encode either image $x^I_i$ or text $x^T_i$, then external prototypes $C^\text{ external}$ can be constructed in the resulting representations space by performing \textit{K}-Means clustering as before. We use PBT to translate the prototypes $C^\text{external}$ to within-modal centroids $C^\text{ external}_{\text{PBT} \to I}$ and $C^\text{ external}_{\text{PBT} \to T}$, then an additional loss term $\mathcal{L}^{\text{ external}}_{\text{Proto}}$ can be added by applying the obtained prototype classifier, taking softmax, then calculating cross-entropy loss in a similar way:
    
    {\begin{equation}\begin{aligned}\label{eq:loss_Proto_external}
        \mathcal{L}^{\text{ external}}_{\text{Proto}}
        = &-(
        \underbrace{\sum_{i=1}^{M}\sum_{k=1}^{K} y^{\text{external}}_{i,k} \log (p^{I\text{, external}}_{i,k})}_{\text{image to external teacher}} \\
        &+ \underbrace{\sum_{i=1}^{M}\sum_{k=1}^{K} y^{\text{external}}_{i,k} \log (p^{T\text{, external}}_{i,k})}_{\text{text to external teacher}} 
        )/2,
    \end{aligned}\end{equation}}

    {where $p^{I\text{, external}}_{i}$ and $p^{T\text{, external}}_{i}$ are the scores obtained by applying the prototype classifier to projected image and text representations, while $y^{\text{external}}_{i}$ indicates the ``ground truth'' of prototype assignment. In practice, we use a pretrained $\text{RoBERTa}_{\text{large}}$
    as the external teacher encoder $E$. During training, the weights of $E$ are frozen so that we can use one epoch of forward pass to cache the teacher features to reduce computation costs. With the external teacher, the loss function of ProtoCLIP becomes:}
    
    {\begin{equation}
        \mathcal{L}_{\text{ProtoCLIP}} =  \mathcal{L}_{\text{Proto}} + \mathcal{L}_{\text{CLIP}} + \mathcal{L}^{\text{ external}}_{\text{Proto}} 
    \end{equation}}

\subsection{Episodic Training}\label{sec:episodic_training}
    Previous clustering-based methods \cite{Caron2018Deep, Asano2020Self, Li2021Prototypical, Alwassel2020Self, Asano2020Labelling, Chen2021Multimodal} update the clusters after an entire training epoch. Such an approach works well on medium-sized ImageNet~\cite{Deng2009ImageNet} dataset since the model can be trained for several hundreds of epochs, resulting in several hundreds of cluster updating. However, CLIP is usually trained for much fewer epochs (e.g., 32~\cite{Cui2022Democratizing}), which makes the frequency of epoch-wise updating insufficient. We propose an \textit{episodic training} strategy. \textit{Episodes} are constructed by randomly choosing $m \ll M$ samples from the entire dataset. Then, 1) feature extraction, 2) prototype updating, and 3) model training are performed sequentially, and then a new episode is then constructed. Episodic training makes prototype updating frequency independent of dataset size $M$, enabling ProtoCLIP to be scaled up to unlimited amounts of training data. To benchmark episodic training-based ProtoCLIP with other models that is trained conventionally, the total number of episode $n_\text{episode}$ is defined as  $n_\text{episode} = n_\text{epoch} \times \frac{M}{m}$. 
    Episode size $m$ is an important hyper-parameter. Smaller $m$ results in higher prototype updating frequency, but too small $m$ increases the sparsity of representations within an episode. In such situations, samples that are assigned to the same prototypes may have different semantics, which decreases the reliability of prototypes.

\section{Experiments}\label{sec:experiments}


    \subsection{Implementation Details}\label{appendix:Implementation Details}

    \subsubsection{Model Architectures}
    Following CLIP~\cite{Radford2021Learning}, we use the modified ResNet-50 backbone as the image encoder, which has three differences compared to the original ResNet-50~\cite{He2016Deep,Li2022Survey}: 1) there are three 3$\times$3 convolutions as ``stem'' instead of a single 7$\times$7 convolution~\cite{He2019Bag}, an average pooling follows the ``stem'' instead of max pooling; 2) the modified ResNet-50 performs antialiased rect-2 blur pooling~\cite{Zhang2019Making}; 3) the final global average pooling layer is replaced with a multi-head self attention~\cite{Vaswani2017Attention, Radford2021Learning}-based pooling. 
    We unitize Transformer~\cite{Vaswani2017Attention, Radford2021Learning} as the text encoder, which consists of 12 layers, 8 attention heads, and a width of 512. The max sequence length is set to 76.
    For image and text projection heads, we use the same architecture as SwAV~\cite{Caron2020Unsupervised}, which is a 2-layers MLP with ReLU activation, 2048 hidden units and 128 output units. Other hyperparameters are summarized in Table~\ref{tab:ProtoCLIP Hyperparameters}.
    
    \subsubsection{Training Configurations} 
    ProtoCLIP is implemented on PyTorch-based OpenCLIP~\cite{Ilharco2021OpenCLIP} codebase. We employ automatic mixed-precision~\cite{Micikevicius2018Mixed} to reduce the training cost. Same as CLIP~\cite{Radford2021Learning}, we use the Adam optimizer~\cite{Kingma2015Adam} with decoupled weight decay regularization~\cite{Loshchilov2019Decoupled}. Gradients are clipped with a maximum norm of 1e5 to prevent model collapse. Learnable temperatures ($\tau_\text{CLIP}$, $\tau_\text{Proto}$) are initialized with 0.07 and clipped by 100 following CLIP~\cite{Radford2021Learning}. Weight decay is not applied to these temperatures. Warm-up and cosine learning rate scheduler~\cite{Loshchilov2017SGDR} are adopted. Same as XDC~\cite{Alwassel2020Self}, we employ early stop to prevent over-fitting. The learning rate of the image encoder is reduced to zero at 16 of 32 epochs, then locked-image tuning~\cite{Zhai2021LiT} is performed for the rest of 16 epochs.
    
    \subsubsection{Prototype Construction} 
    We adopt Faiss~\cite{Johnson2019Billion} implemented \textit{K}-Means for clustering. We cluster the 128-dimensional projected representations (i.e., $h^I, h^T$) of 200,000 samples in each episode to \textit{K}=20,000 clusters and use the resulting cluster centroids as prototypes. \textit{K}-Means is optimized for 20 iterations, which we found it sufficient for convergence. We use a pretrained $\text{RoBERTa}_{\text{large}}$\footnote{ \url{https://pytorch.org/hub/pytorch_fairseq_roberta}} as the external teacher. We extract $\text{RoBERTa}_{\text{large}}$ representations off-line to speed-up ProtoCLIP training, and reduce the representation dimension from 1024 to 64 by PCA to save memory cost.

    \begin{table*}
        \caption{ProtoCLIP Hyper-parameters.}
        \label{tab:ProtoCLIP Hyperparameters}
        \centering
        \begin{tabular}{ccc}
        
        \toprule
        {Section}                                   & {Hyperparameter}                                                  & {Value}                                   \\
        
        \midrule
        \multirow{3}{*}{Episodic Training}        & Batch size                                                      & 512                                     \\
                                                  & Episode size                                                    & 200,000                                 \\
                                                  & Warm-up Episodes                                                 & 40                                   \\
        
        \midrule
        \multirow{2}{*}{Prototype   Construction} & Number of clusters in \textit{K}-Means                                           & 20,000                                  \\
                                                  & \textit{K}-means Iterations                                      & 20                                      \\
        
        \midrule
        \multirow{5}{*}{Optimization}             & Optimizer                                                       & Adam                                    \\
                                                  & Adam $\beta_1$, $\beta_2$, $\epsilon$                          & 0.9, 0.999, 1e-8                        \\
                                                  & Learning Rate                                                   & 5e-4, cosine decay                      \\
                                                  & Weight decay                                                    & 0.5                                     \\
                                                  & Maximum gradient norm                                           & 1e5                                     \\
        
        \midrule
        \multirow{9}{*}{Model Architectures}      & Image Encoder                                                   & ResNet-50/ResNet-101                      \\
                                                  & Image Resolution                                                & 224$\times$224                          \\
                                                  & Text Encoder                                                    & Transformer                 \\
                                                  & Text vocabulary size                                            & 49408                                   \\
                                                  & Initial and maximum temperature   ($\tau_\text{CLIP}$, $\tau_\text{Proto}$) & 0.07, 100                                    \\                                                  & Representation dimension   ($d_z$)                              & 1024                                    \\
                                                  & Projected Representation   dimension ($d_h$)                    & 128                                     \\
                                                  & External Teacher                               & $\text{RoBERTa}_{\text{large}}$ \\  
        \bottomrule
        \end{tabular}%
    \end{table*}
    
    \subsection{Pretraining Data}
        \subsubsection{Conceptual Captions}~\cite{Sharma2018Conceptual} it is an webly collected high-quality image-text dataset consist of 3,318,333 sample pairs. The dataset was made public\footnote{ \url{https://github.com/google-research-datasets/conceptual-captions}} by Google in 2018. Unfortunately, the number of accessible images keeps drooping due to expired image links. This issue is raised by several recent works in the field of VLP~\cite{Fuerst2021CLOOB, Li2021Supervision, Ilharco2021OpenCLIP}. In this work, since we can only collect 2,643,718 images, we randomly sample a 2,500,000 subset (75\% of full CC3M) from them to train our ProtoCLIP. Considering the dropping accessibility of image links in Conceptual Captions, we call for the use of this dataset size (2.5M) in future benchmarking for better comparability.
        
        {\subsubsection{YFCC} it was created by filtering YFCC100M~\cite{Thomee2016YFCC100M} for images which contain natural language descriptions and/or titles in English. We used the ``YFCC15M-v1'' filtered by OpenAI~\cite{Radford2021Learning}. We obtained about 14M out of the total 15M samples due to broken links, and used 14,000,000 of them to train the ProtoCLIP.}

        \subsubsection{Data Augmentations}
        Recent advances in VLP \cite{Mu2021SLIP, Li2021Supervision, Yao2021FILIP} have shown that applying random data augmentations can be beneficial. However, we found that common data augmentation strategies used in image SSL is too aggressive in the VLP scenario. As shown in Figure~\ref{fig:augmentations}, standard SimCLR~\cite{Chen2020Simple} augmentations have a higher chance of changing semantics when it is applied to non-iconic images of Conceptual Captions dataset\footnote{Several recent works of image SSL have also pointed out that applying SimCLR augmentations on non-iconic scenes images is not optimal. For more details please see ORL~\cite{Xie2021Unsupervised} and UniVIP~\cite{Li2022UniVIP}.}. Such semantic inconsistency poses extra difficulty to image-text representation alignment. To this end, we design a lighter data augmentation to train ProtoCLIP by making two modifications to the SimCLR augmentations parameters: 1) images are randomly resized and cropped with a scale range of 50\% to 100\%  instead of 8\% to 100\%; 2) probability of applying color jittering is reduced from 0.8 to 0.2. As Figure~\ref{fig:augmentations}, such data augmentation maintains higher semantic consistency than that of SimCLR augmentations.
        
        We note that, with applied random data augmentations, our proposed episodic training strategy and PBT can implicitly create additional contrastive supervision for image representations. Recall that episodic training consists of three steps including 1) feature extraction, 2) prototype construction, and 3) model training. Since the first and the third step is performed independently, different augmentations are drawn and applied to the same image. During the model training step, the representation of an image is pushed to the assigned and translated centroid of its another view built in the feature extraction step, leading to an additional contrastive supervision.
        
        Such implicit contrast shares some similarities with SwAV and DeepCluster-v2 that learn visual representations by ``contrasting cluster assignments''~\cite{Caron2020Unsupervised}. However, they use the cluster assignment to set up within-modal supervision, while the implicit contrast of ProtoCLIP is done through the text representation space. Recent SLIP~\cite{Mu2021SLIP} and DeCLIP~\cite{Li2021Supervision} also applied data augmentation-based contrast to boost VLP performance. However, they contrasted image representations explicitly by forward additional views of images in each training step, which leads to a significantly expanded memory footprint and decreased maximum allowed batch size.\footnote{Many recent studies have proved that sufficiently large batch size is crucial for contrastive learning~\cite{Ash2022Investigating}.} In our ProtoCLIP, two views for the implicit contrast are built separately during feature extraction and model training. Although it leads to additional time consumption, the maximum allowed batch size is not affected. 
        
    \begin{figure*}
        \centering
        \includegraphics[width=0.95\linewidth]{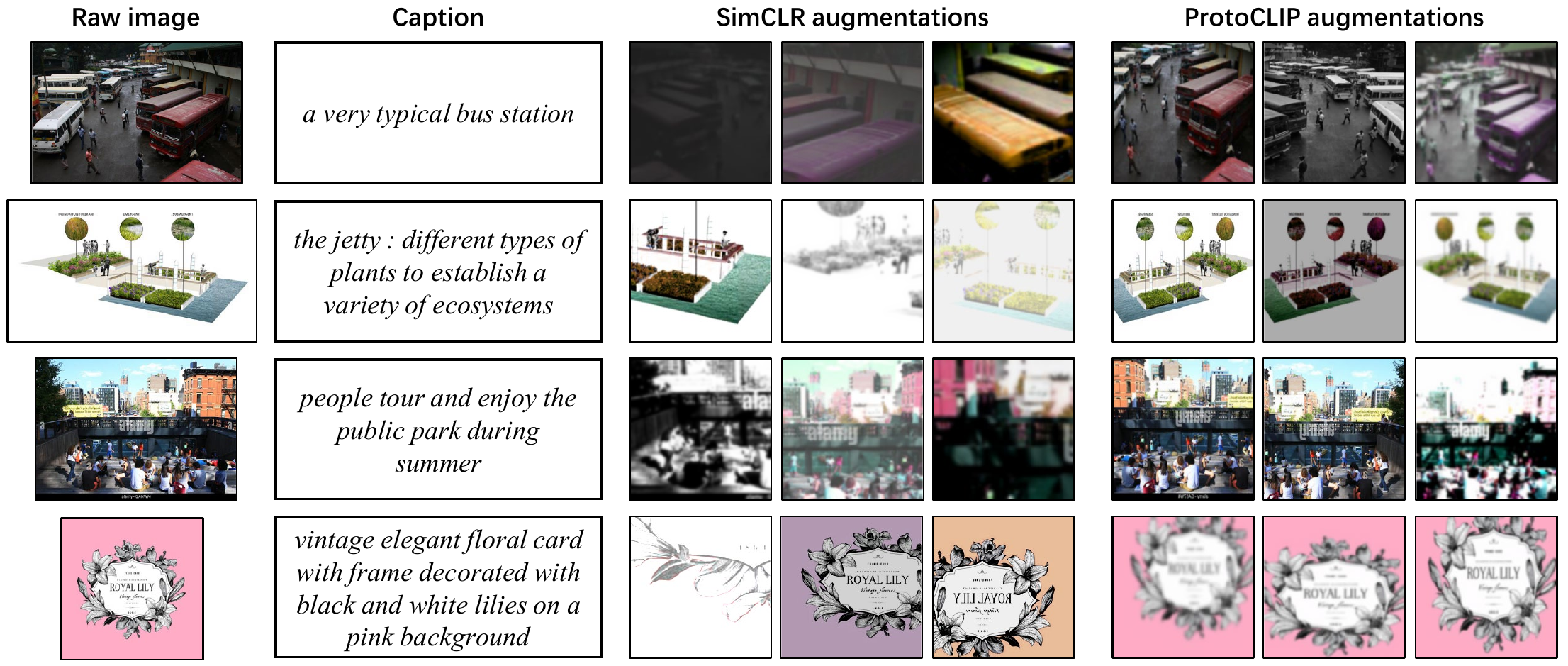}
        \caption{Visualization of different data augmentations. ProtoCLIP augmentations maintain higher semantic consistency on non-iconic images in Conceptual Captions.}
        \label{fig:augmentations}
    \end{figure*}

\subsection{Evaluation Metrics }\label{appendix: CC exp details}
\subsubsection{Zero-shot Classification}
    We use the 1024-dimensional L2-normalized representations (i.e., $z^I, z^T$) extracted by image and text encoders to perform zero-shot classification. Class names and prompt templates are consistent with CLIP~\cite{Radford2021Learning} in spite of minor explanations to some classes, e.g., ``kite''$\to$``kite (bird of prey)'' are added following CLOOB~\cite{Fuerst2021CLOOB}. {A total of 14 datasets are adopted: including ImageNet~\cite{Deng2009ImageNet}, DTD~\cite{Cimpoi2014Describing}, Food101~\cite{Bossard2014Food}, Oxford-IIIT Pet~\cite{Parkhi2012Cats}, RenderedSST2~\cite{Radford2021Learning}, Birdsnap~\cite{Berg2014Birdsnap}, Country211~\cite{Radford2021Learning}, Flowers102~\cite{Nilsback2008Automated}, GTSRB~\cite{Stallkamp2011German}, UCF101~\cite{Soomro2012UCF101}, Stanford Cars~\cite{Krause20133D}, CIFAR10, CIFAR100 and STL10}, whose details are summarized in Table~\ref{tab:zeroshot_datasets}. Similar to the Conceptual Captions dataset, the Birdsnap dataset also faces the problem of link expiration. Same as CLIP~\cite{Radford2021Learning} and CLOOB~\cite{Fuerst2021CLOOB}, we use the resources that are available online at the time of writing. 

    \begin{table*}[h]
    \centering
    \caption{Dataset used in zero-shot classification evaluation.}
    \label{tab:zeroshot_datasets}
        \begin{tabular}{cccl}
        \toprule
        Dataset       & Classes & Testset Size & \multicolumn{1}{c}{Description}                \\
        \midrule
        ImageNet      & 1,000   & 50,000        & 1000 categories of objects                     \\
        DTD           & 47      & 1,880         & 47 categories of texture patches                     \\
        Food101       & 101 & 25,250            &  101 categories of food dishes                     \\
        Oxford-IIIT Pet & 37    & 3,669         &  37 breeds of cats and dogs                \\
        RenderedSST2  & 2       & 1,821         &  2 classes of positive or negative movie reviews rendered as text       \\
        Birdsnap      & 500     & 1,855         & 500 categories of North American bird species  \\
        Country211    & 211     & 21,100        & 211 countries represented by geo-tagged images \\
        Flowers102    & 102     & 6,149         & 102 species of common UK flowers               \\
        GTSRB         & 43      & 12,630        & 43 categories of German traffic signs          \\
        UCF101        & 101     & 11,213        & 101 categories of human actions using the middle frame of each clip  \\
        Stanford Cars & 196     & 8,041         & 196 categories of cars (make, model, and year)\\
        CIFAR10       & 10      & 10,000        & 10 categories of animals and   vehicles        \\
        CIFAR100      & 100     & 10,000        & 100 categories of animals, vehicles, plants, objects, scenes, people \\
        STL10         & 10      & 8,000         & 10 categories of animals and   vehicles        \\
        \bottomrule
        \end{tabular}
    \end{table*}
   
\subsubsection{Linear Probing}
    Frozen 1024-dimensional image representations ($z^I$) before normalization are used for linear probing. For small-scale CIFAR10, CIFAR100, and STL10, we train a logistic regression classifier using scikit-learn’s L-BFGS implementation, with a maximum of 1,000 iterations following CLIP~\cite{Radford2021Learning}. For larger ImageNet dataset, we adopt PyTorch-based SGD optimization following MoCo~\cite{He2020Momentum}, SwAV~\cite{Caron2020Unsupervised} and SLIP~\cite{Mu2021SLIP} to utilize GPU efficiency. Specifically, we train a linear classifier for 100 epochs with a batch size of 256, a learning rate of 0.3, and a weight decay of 1e-6. SGD optimizer with a momentum of 0.9 and cosine learning rate scheduler are applied.

{
\subsubsection{K-NN Classification}
    Following DINO~\cite{Caron2021Emerging}, we applied K-NN classification to evaluate the quality of image representations. We build K-NN classifiers based on the training set representations, then measure the top-1 classification accuracy on testing set. K-NN is less sensitive to hyper-parameters compared to linear probing, and we set K=20 for all datasets~\cite{Caron2021Emerging}.}

\subsubsection{Image-text Retrieval}
    Image-text retrieval task consists of image to text retrieval and text to image retrieval. The performance is evaluated on MS-COCO~\cite{Chen2015Microsoft} benchmark under the zero-shot setting (i.e., without fine-tuning). The dot-similarity of L2-normalized 1024-dimensional image and text representations ($z^I, z^T$) are used for ranking. We report recall@1, recall@5 and recall@10 and their average as mean recall.

{
\subsubsection{Object Detection}
    The image encoder is transferred to perform object detection. We adopted Mask R-CNN detector~\cite{He2020Mask}, and fine-tune the pretrained encoder on MS-COCO dataset for 12 epochs following DenseCLIP~\cite{Rao2021DenseCLIP}.}

\subsection{Ablation Study {of ProtoCLIP Hyper-parameters}}\label{sec:ablation_study}

    This section validates the impact of the hyper-parameters of ProtoCLIP. A one-million subset of the Conceptual Captions (CC)~\cite{Sharma2018Conceptual} dataset is used. To avoid testset hyper-parameter tuning, CIFAR10, CIFAR100 and STL10 dataset are adopted here for validation. Benchmarks on other downstream datasets will be reported later in Section~\ref{sec:CC_benchmark} and Section~\ref{sec:YFCC_benchmark}. Total training amount here (episode size$\times n_\text{episode}$) is set equivalent to 20 epochs. 
    The default setting of the ProtoCLIP includes episode size $m=0.2$M, no soft target, 10 images per prototype, no external teacher, and no data augmentations. \textit{K}-Means is performed with a max iteration limit of 20 steps, which we found sufficient to converge. 

\subsubsection{Episode Size} 
    As illustrated in Section~\ref{sec:episodic_training}, there exists a trade-off between prototype reliability and updating frequency. Here, we try to find an optimal episode size that can satisfy both sides by training ProtoCLIP (without $\mathcal{L}_{\text{CLIP}}$) using different episode size. As shown in Figure~\ref{fig:ablation}(a), an episode size of 0.2M yields the best performance. The rightmost bar in red (episode size=1M) is equivalent to update the cluster after one entire training epoch as done in previous methods~\cite{Caron2018Deep, Asano2020Self, Li2021Prototypical, Alwassel2020Self, Asano2020Labelling, Chen2021Multimodal}. With the best value of episode size, our episodic training strategy leads to a +2.86\% improvement.
    
\subsubsection{Target Temperature}
    Next, we turn to select the best target temperature $\tau_{\text{y}}$. Although a higher value of $\tau_{\text{y}}$ transfers structural relation knowledge, too large $\tau_{\text{y}}$ makes target scores to be over-smoothed. Figure~\ref{fig:ablation}(b) shows that $\tau_{\text{y}}$=0.01 achieves the best performance. Compared to the one-hot label (hardmax, the leftmost bar in red) used in previous clustering-based SSL approaches~\cite{Caron2018Deep, Asano2020Self, Alwassel2020Self, Asano2020Labelling, Li2021Prototypical, Zhang2021Supporting}, learning from soft target brings +1.58\% improvement.
    
\subsubsection{Number of Images per Prototype} 
    Clustering-based SSL for ImageNet pretraining often sets the total number of clusters to be several thousands (e.g., $K=3000$ for SwAV~\cite{Caron2020Unsupervised}). We found that with uncurated image-text dataset, this hyperparameter should be chosen more conservatively. The reason is that uncurated image-text dataset contains much more concepts than curated ones~\cite{Yang2022Unified}. Lower $K$ increases the noise within each cluster. We train our model with $\mathcal{L}_{\text{Proto}}$ + $\mathcal{L}_{\text{CLIP}}$. Figure~\ref{fig:ablation}(c) shows that 10 images per prototype (i.e. $K=20$k for an episode size of 0.2M) yield the best performance.
    
\subsubsection{External Teacher}
    Finally, we compare different external teachers. We consider the text encoder of pretrained CLIP (ViT/B-32)~\cite{Radford2021Learning} and the pretrained RoBERTa~\cite{Liu2019RoBERTa}. Figure~\ref{fig:ablation}(d) shows that both of these two external teacher benefit ProtoCLIP, while RoBERTa brings more improvement.

\begin{figure*}
    \centering
    \includegraphics[width=0.85\linewidth]{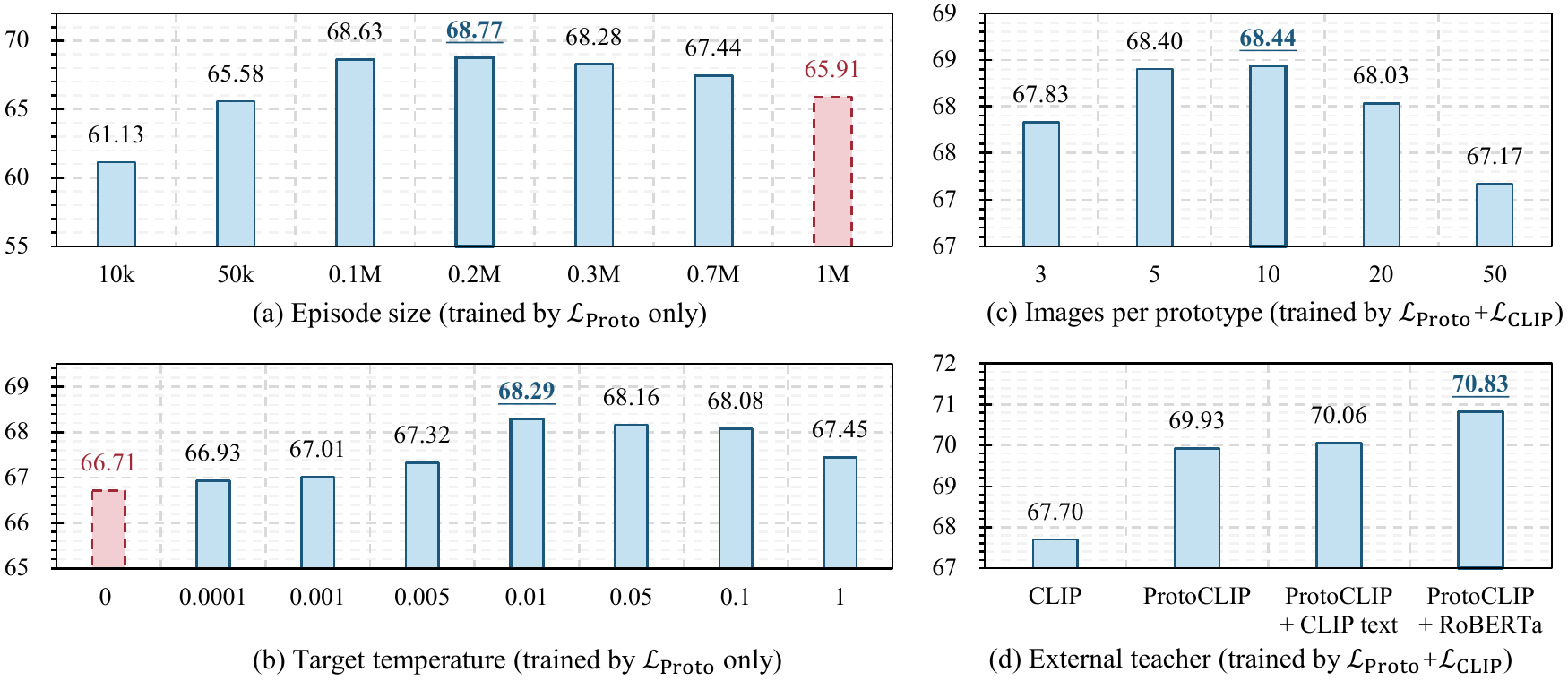}
    \caption{ProtoCLIP ablation experiments on Conceptual Captions 1M data (20 epoch). We report the average linear probing accuracies (\%) of CIFAR10, CIFAR100, and STL10. Detailed results are given in Appendix~\ref{appendix: CC exp details}.}
    \label{fig:ablation}
\end{figure*}

\subsection{Conceptual Captions Pretraining}\label{sec:CC_benchmark}

        
    \begin{figure}
    \centering
    \includegraphics[width=\linewidth]{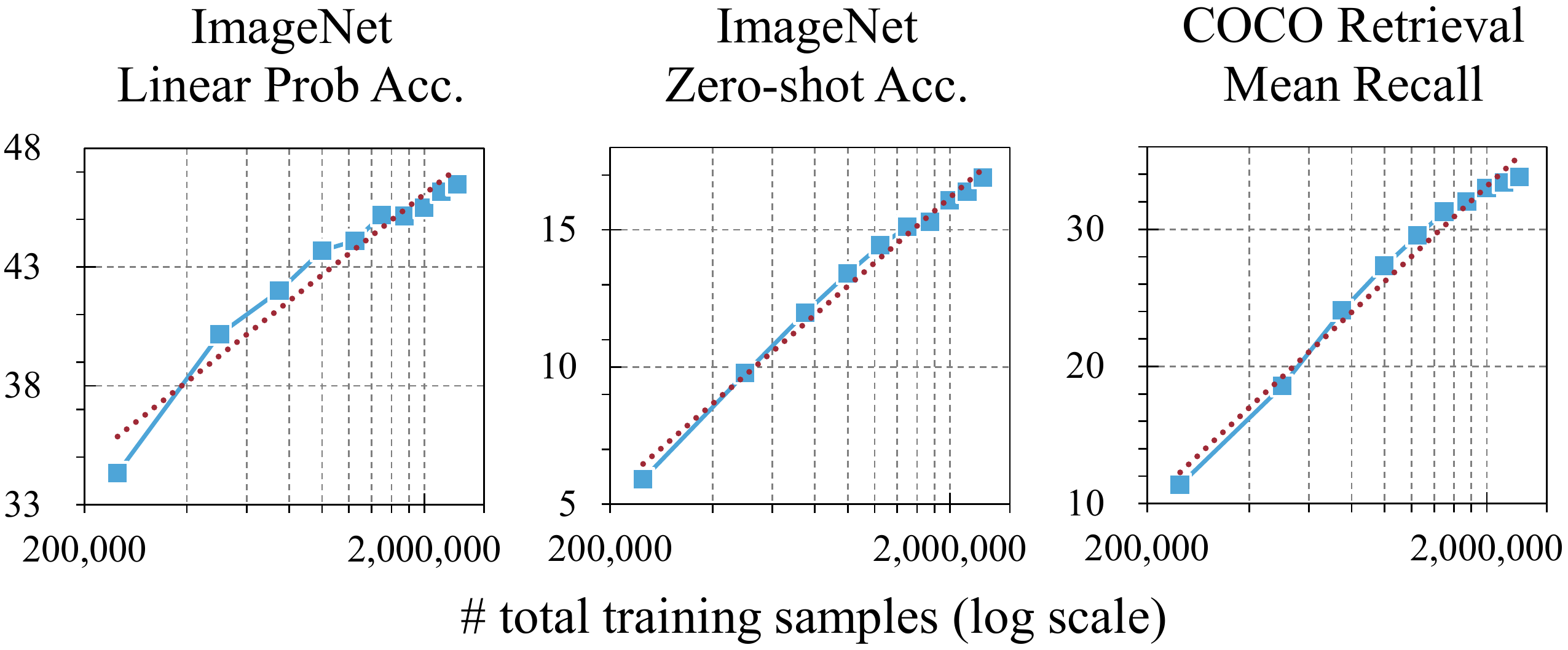}
            \caption{\footnotesize  {CLIP performance scales steadily with dataset size.}}
            \label{fig:dataset_scaling_trends}
    \vspace{-0.3cm}
    \end{figure}
    
    {With selected hyper-parameters, we now train the ProtoCLIP on full CC data. 
    The original CC dataset~\cite{Sharma2018Conceptual} (collected in 2018) contains over 3.3M samples. Unfortunately, due to broken links, an increasing number of images become inaccessible. To benefit future benchmarking, we use a total of 2,500,000 samples (CC2.5M) from CC to train our model. 
    Such size is much small than that of the original CLIP~\cite{Radford2021Learning}. However, as in Figure~\ref{fig:dataset_scaling_trends}, we train CLIP with different dataset sizes and found the downstream performance of the CLIP model (blue) scales up steadily (near-logarithmically, as noted by red dotted lines) with dataset size. This is also demonstrated by Ilharco et al.~\cite{Ilharco2021OpenCLIP}. Therefore, the dataset size of CC2.5M is already able to reflect the effectiveness of VLP models accurately.}
    We continue to adopt ResNet-50~\cite{He2016Deep} and transformer~\cite{Vaswani2017Attention} as image and text encoders. With a single-node 4$\times$2080Ti machine, training a ProtoCLIP for 32 epochs on full CC2.5M takes approximately 64 hours. 
    
    \subsubsection{Downstream Evaluations} We perform linear probing and zero-shot classification on ImageNet, CIFAR and STL. {We note that CLIP models are usually evaluated on a more diverse downstream datasets. However, since the CC dataset cannot provide sufficient coverage for open-vocabulary visual concepts~\cite{Yang2022Unified}, the downstream performance on these diverse datasets is extremely low (e.g., $<10\%$) and the performance differences between different methods are marginal, making it hard to draw safe conclusion based on the metrics. Instead, we leave comprehensive evaluations on diverse datasets to Section~\ref{sec:YFCC_benchmark} when the models are trained with larger YFCC dataset.} Moreover, mean recall of MS-COCO~\cite{Chen2015Microsoft} cross-modal retrieval is also reported to evaluate representation alignment. 
    
    \subsubsection{Effectiveness of Each Component} We first validate the effectiveness of each ProtoCLIP component on CC2.5M. We train ProtoCLIP on CC2.5M for 8 epochs, and compare its zero-shot classification and linear probing performance {with CLIP and} ablations of ProtoCLIP. Classification accuracy on ImageNet and averaged accuracy on CIFAR10, CIFAR100, and STL10 are reported. We first remove the external teacher RoBERTa, then respectively ablate 1) PBT, 2) soft target, 3) \textit{K}-Means optimizing, and 4) data augmentation. As in Table~\ref{tab:CC2.5M_benchmark:a}, full ProtoCLIP achieve the best performance overall. Every other comparison yields degenerated performance, showing the effectiveness of each component. For ImageNet linear probing accuracy, introducing PBT brings +1.83\% improvement, while introducing an external teacher brings +1.76\% improvement. {All these ablations of ProtoCLIP outperform CLIP baseline on most metrics.}
    
    {\subsubsection{Benchmarking ProtoCLIP} {Next, we benchmark ProtoCLIP by training it on CC2.5M for standard 32 epochs, and further half the epochs of ProtoCLIP or double the epochs of CLIP baseline to compare training efficiency. Table~\ref{tab:CC2.5M_benchmark:b} summarizes the main results. With the same 32 training epochs (line 2 vs. 3), ProtoCLIP outperforms CLIP by +5.81\% on ImageNet linear probing and +2.01\% on ImageNet zero-shot classification.  As ProtoCLIP takes additional steps during training (feature extraction, \textit{k}-Means clustering, etc.), we also report the absolute training time via the number of ``Relative Epoch'' (Rel.) in comparison with one standard CLIP epoch.  Comparing the first two rows, using 67\% of training time (ProtoCLIP 21.5 Rel. / CLIP 32.0 Rel.), ProtoCLIP outperform CLIP on ImageNet zero-shot Top-1, Top-5, linear probing, and matched the performance of CLIP in retrieval. Similarly, comparing the third and fourth rows (ProtoCLIP 43.1 Rel. vs. CLIP 64.0 Rel.), similar observations can also be obtained. These comparisons demonstrated the efficiency of ProtoCLIP.   
    Interestingly, ProtoCLIP also shows promising potential for improving data efficiency, even though it is not our main design objective. Using 2.5M data, ProtoCLIP achieves 20.39\% ImageNet zero-shot classification accuracy, marginally exceeding that of a CLIP model (20.33\%, reported in \cite{Fuerst2021CLOOB}) which uses additional ~0.4M data. At the same time, such a result of ProtoCLIP is achieved by using around 2/3 of training time (ProtoCLIP 21.5 Rel. vs. CLIP 31 epoch).} 


\begin{table}[t]\centering
\caption{ProtoCLIP ablation experiment on Conceptual Captions 2.5M data (8 epoch).}
\label{tab:CC2.5M_benchmark:a}
    \resizebox{0.9\linewidth}{!}{
    \begin{tabular}{llcccc}
    \multicolumn{2}{c}{Method} & 
        \rotatebox{90}{\begin{tabular}[c]{@{}l@{}}ImageNet\\ zero-shot\end{tabular}} & 
        \rotatebox{90}{\begin{tabular}[c]{@{}l@{}}ImageNet\\ linear\end{tabular}} & 
        \rotatebox{90}{\begin{tabular}[c]{@{}l@{}}CIFAR \& STL \\ zero-shot Avg.\end{tabular}} & 
        \rotatebox{90}{\begin{tabular}[c]{@{}l@{}}CIFAR \& STL \\ linear Avg. \end{tabular}} \\
        \toprule
            \multicolumn{2}{c}{{CLIP}} 
            & {9.89} 
            & {41.30} 
            & {38.77} 
            & {67.32} \\
        \rowcolor[HTML]{DAE8FC} 
            \multicolumn{2}{c}{\textbf{ProtoCLIP}} 
            & \textbf{\uline{11.96}} 
            & \textbf{\uline{46.55}} 
            & 42.74 & \uline{\textbf{70.96}} \\
        \rowcolor[HTML]{DAE8FC} 
        \multicolumn{2}{c}{ProtoCLIP w/o RoBERTa} & 11.91 &44.76 & \textbf{\uline{42.81}} & 69.45 \\
         & - w/o PBT & 11.23            &42.93 & 42.32 & 68.89 \\
         & - w/o soft target & 11.28    &44.22 & 42.66 & 69.18 \\
         & - w/o \textit{K}-means & 11.62         &44.27 & 38.67 & 67.22 \\
         & - w/o augmentation & 11.17   &44.39 & 38.67 & 68.75 \\
    \bottomrule
    \end{tabular}
    }
\end{table}

\begin{table}[t]
\caption{Conceptual Captions pretraining benchmarks. The ``$\vartriangleright$'' indicates results reported by corresponding papers.}
\label{tab:CC2.5M_benchmark:b}
    \resizebox{\linewidth}{!}{\begin{tabular}{ccrlrrrr}
    \begin{tabular}[c]{@{}c@{}}Batch\\ Size\end{tabular} & Data & {Epoch (Rel.)} & Method & 
        \rotatebox{90}{\begin{tabular}[c]{@{}l@{}}ImageNet \\ zero-shot top-1\end{tabular}} & 
        \rotatebox{90}{\begin{tabular}[c]{@{}l@{}}ImageNet \\ zero-shot top-5\end{tabular}} & 
        \rotatebox{90}{\begin{tabular}[c]{@{}l@{}}ImageNet \\ linear\end{tabular}} & 
        \rotatebox{90}{\begin{tabular}[c]{@{}l@{}}retrieval \\ mean recall\end{tabular}} \\ 
        \toprule
        \multirow{4}{*}{512} & \multirow{4}{*}{2.5M} 
        & \cellcolor[HTML]{DAE8FC} {16 (21.5)}
        & \cellcolor[HTML]{DAE8FC} {\textbf{ProtoCLIP} }
            & \cellcolor[HTML]{DAE8FC} {20.39}
            & \cellcolor[HTML]{DAE8FC} {40.02}
            & \cellcolor[HTML]{DAE8FC} {50.47}
            & \cellcolor[HTML]{DAE8FC} {36.01}   \\ 
        &  & {32 (32.0)} & {CLIP}
            & {19.46}
            & {38.42}
            & {49.41}
            & {36.48} \\
        &  & \cellcolor[HTML]{DAE8FC} {32 (43.1)}
        & \cellcolor[HTML]{DAE8FC} {\textbf{ProtoCLIP}}
            & \cellcolor[HTML]{DAE8FC} {21.47}
            & \cellcolor[HTML]{DAE8FC} {40.84}
            & \cellcolor[HTML]{DAE8FC} {55.22}
            & \cellcolor[HTML]{DAE8FC} {35.69}  \\ 
        &  & {64 (64.0)} & {CLIP }
            & {20.34}
            & {39.21}
            & {51.14}
            & {37.61} \\
        \cline{1-3}
        \multirow{2}{*}{512} & \multirow{2}{*}{2.9M} & \multirow{2}{*}{31} & $\vartriangleright$ CLOOB~\cite{Fuerst2021CLOOB} & 23.97 & -  & - & - \\ 
         &  &  & $\vartriangleright$ CLIP & 20.33 & - & - & - \\ \cline{1-3}
        \multirow{2}{*}{1024} & \multirow{2}{*}{3M} & \multirow{2}{*}{32} & $\vartriangleright$ DeCLIP~\cite{Li2021Supervision} & 27.2 & - & - & - \\ 
         &   &   & $\vartriangleright$ CLIP & 20.6 & - & - & - \\ 
    \bottomrule
    \end{tabular}}
\end{table}

\subsection{YFCC Pretraining}\label{sec:YFCC_benchmark}
    We train the ProtoCLIP on the YFCC-100M~\cite{Thomee2016YFCC100M} subset YFCC-15M filtered by OpenAI~\cite{Radford2021Learning}, which consists about 15 million of image-text pairs. We used the ResNet-50~\cite{He2016Deep} as well as a larger ResNet-101 backbone. The CLIP checkpoints (YFCC-15M, 32 epochs) released by OpenCLIP~\cite{Ilharco2021OpenCLIP} are adopted as the baseline. 
       
    Downstream performance is summarized in Table~\ref{tab:YFCC_benchmark_summary}. Impressively, the performance of ProtoCLIP matched CLIP witharound 1/3 pretraining time costs, which demonstrates that ProtoCLIP significantly improves the representation learning efficiency. Zero-shot accuracies for each dataset can be found in Table~\ref{tab:YFCC_full_zero_shot}, which shows that ProtoCLIP outperforms CLIP on 9 of 14 datasets. Compared to linear probing, ProtoCLIP has more advantages in K-NN classification. Linear probing measures the linear separability of the image representations, while K-NN measures whether semantically similar samples are clustered correctly. We argue that ProtoCLIP's advantage in K-NN classification is brought by the prototypical contrast and efficient representation grouping. Clustering evaluation in Section~\ref{sec:Clustering Evaluation} further proves this property. The quality of ProtoCLIP image representations is confirmed by transfer learning for object detection: 8 epoch (10.8 Rel.) ProtoCLIP matched the result of 32 epoch (32.0 Rel.) CLIP. Interestingly, we found that ProtoCLIP yields notable improvement in retrieval, which seems to be a contradiction of the observation on Conceptual Captions pretraining. We leave the investigation of this phenomenon to future work.
    
    \begin{table*}[t]
    \caption{\textbf{Summary of YFCC pretraining results.} Performance comparison of zero-shot / linear / K-NN classification and COCO retrieval. ``IN'': ImageNet, ``14/9 Avg.'': averaged classification accuracy across datasets. The performance of ProtoCLIP matched CLIP with around 1/3 pretraining time costs.}
    \label{tab:YFCC_benchmark_summary}
    \centering
    \begin{tabular}{cccrccccccc}
    \toprule
     &
       &
       &
       &
      \multicolumn{2}{c}{\textbf{Zero-shot}} &
      \multicolumn{2}{c}{\textbf{Linear}} &
      \multicolumn{2}{c}{\textbf{K-NN}} &
      \textbf{Retrieval} \\
    \multirow{-2}{*}{\textbf{Arch.}} &
      \multirow{-2}{*}{\textbf{Model}} &
      \multirow{-2}{*}{\textbf{Data}} &
      \multirow{-2}{*}{\textbf{{Epcoh (Rel.)}}} &
      IN &
      14 Avg. &
      IN &
      9    Avg. &
      IN &
      9    Avg. &
      Mean Recall \\ \midrule
     &
      CLIP &
      15M &
      32 (32.0) &
      32.7 &
      31.1 &
      61.5 &
      65.8 &
      56.0 &
      57.8 &
      40.9 \\
    \multirow{-2}{*}{RN50} &
      ProtoCLIP &
      14M &
      8 (10.8) &
      32.0 &
      31.9 &
      62.1 &
      65.4 &
      56.7 &
      58.3 &
      42.7 \\
    \multicolumn{1}{l}{} &
      \multicolumn{3}{c}{{\color[HTML]{808080} }} &
      \cellcolor[HTML]{DAE8FC}{\color[HTML]{FF0000} -0.7} &
      \cellcolor[HTML]{DAE8FC}{\color[HTML]{00B050} +0.8} &
      \cellcolor[HTML]{DAE8FC}{\color[HTML]{00B050} +0.6} &
      \cellcolor[HTML]{DAE8FC}{\color[HTML]{FF0000} -0.3} &
      \cellcolor[HTML]{DAE8FC}{\color[HTML]{00B050} +0.8} &
      \cellcolor[HTML]{DAE8FC}{\color[HTML]{00B050} +0.5} &
      \cellcolor[HTML]{DAE8FC}{\color[HTML]{00B050} +1.8} \\ \midrule
     &
      CLIP &
      15M &
      32 (32.0) &
      34.8 &
      32.9 &
      63.1 &
      66.2 &
      57.9 &
      59.0 &
      43.2 \\
    \multirow{-2}{*}{RN101} &
      ProtoCLIP &
      14M &
      8 (10.8) &
      33.8 &
      33.0 &
      62.9 &
      65.4 &
      58.0 &
      59.0 &
      44.7 \\
    \multicolumn{1}{l}{} &
      \multicolumn{3}{c}{{\color[HTML]{808080} }} &
      \cellcolor[HTML]{DAE8FC}{\color[HTML]{FF0000} -1.0} &
      \cellcolor[HTML]{DAE8FC}{\color[HTML]{00B050} +0.1} &
      \cellcolor[HTML]{DAE8FC}{\color[HTML]{FF0000} -0.2} &
      \cellcolor[HTML]{DAE8FC}{\color[HTML]{FF0000} -0.8} &
      \cellcolor[HTML]{DAE8FC}{\color[HTML]{00B050} +0.2} &
      \cellcolor[HTML]{DAE8FC}{\color[HTML]{00B050} +0.1} &
      \cellcolor[HTML]{DAE8FC}{\color[HTML]{00B050} +1.5} \\ \bottomrule
    \end{tabular}
    \end{table*}

    Here we also present full results of zero-shot classification (Table~\ref{tab:YFCC_full_zero_shot}), linear probing and K-NN classification (Table~\ref{tab:YFCC_full_linear}), zero-shot image-text retrieval (Table~\ref{tab:YFCC_full_retrieval}) and MS-COCO object detection (Table~\ref{tab:YFCC_full_COCO}). These results are the detailed result in Table~\ref{tab:YFCC_full_zero_shot}.

    \begin{table*}[h]
    \centering
    \caption{{\textbf{Zero-shot classification} evaluation results of YFCC pretraining.}}
    \label{tab:YFCC_full_zero_shot}
    \resizebox{\textwidth}{!}{%
    \begin{tabular}{ccccccccccccccccccc}
    Arch. & Method & Data & {Epcoh (Rel.)}
    & \rotatebox{90}{\begin{tabular}[c]{@{}l@{}}Birdsnap \end{tabular}} 
    & \rotatebox{90}{\begin{tabular}[c]{@{}l@{}}CIFAR10 \end{tabular}} 
    & \rotatebox{90}{\begin{tabular}[c]{@{}l@{}}CIFAR100 \end{tabular}} 
    & \rotatebox{90}{\begin{tabular}[c]{@{}l@{}}Country211 \end{tabular}} 
    & \rotatebox{90}{\begin{tabular}[c]{@{}l@{}}DTD \end{tabular}} 
    & \rotatebox{90}{\begin{tabular}[c]{@{}l@{}}Flowers102 \end{tabular}} 
    & \rotatebox{90}{\begin{tabular}[c]{@{}l@{}}Food101 \end{tabular}} 
    & \rotatebox{90}{\begin{tabular}[c]{@{}l@{}}GTSRB \end{tabular}} 
    & \rotatebox{90}{\begin{tabular}[c]{@{}l@{}}OxfordIIITPet \end{tabular}} 
    & \rotatebox{90}{\begin{tabular}[c]{@{}l@{}}RenderedSST2 \end{tabular}} 
    & \rotatebox{90}{\begin{tabular}[c]{@{}l@{}}StanfordCars \end{tabular}} 
    & \rotatebox{90}{\begin{tabular}[c]{@{}l@{}}STL10 \end{tabular}} 
    & \rotatebox{90}{\begin{tabular}[c]{@{}l@{}}UCF101 \end{tabular}} 
    & \rotatebox{90}{\begin{tabular}[c]{@{}l@{}}ImageNet \end{tabular}} 
    & \rotatebox{90}{\begin{tabular}[c]{@{}l@{}}14 Dataset Avg.\end{tabular}}  \\
    \toprule
    \multirow{2}{*}{RN50} & CLIP & 15M & 32 (32.0) & 21.81 &	49.12 &	20.32 &	6.34 &	17.55    & 50.24 &	42.80 &	9.57 &	27.45 &	49.92 &	3.99 &	79.14 &	24.32 &	32.72 &	31.09 \\
     & ProtoCLIP & 14M & 8 (10.8) &  19.92 &	53.95 &	24.60 &	7.14 &	20.00 &	50.51 &	38.61 &	    7.19 &	23.88 &	50.08 &	4.68 &	86.13 &	27.84 &	32.02 &	31.90 \\
      &  &  &  &  -1.88 & \textbf{+4.83} &	\textbf{+4.28} &	\textbf{+0.80} &	\textbf{+2.45} &	\textbf{+0.28} &	-4.19 &	-2.38 &	-3.57 &	\textbf{+0.16} & 	\textbf{+0.68} & 	\textbf{+6.99} &	\textbf{+3.52} &	-0.70 &	\textbf{+0.80} \\
     \midrule
    \multirow{2}{*}{RN101} & CLIP (32.0) & 15M & 32 & 22.94 &	52.99 &	22.94 &	6.82 &	18.24 &	50.40 &	44.18 &	9.41 &	30.25 &	49.92 &	3.69 &	86.68 &	27.44 &	34.84 &	32.91  \\
     & ProtoCLIP & 14M & 8 (10.8) &  19.49 &	57.23 &	26.50 &	7.92 &	19.68 &	52.02 &	39.60 &	7.34 &	25.78 &	50.03 &	5.04 &	88.75 &	28.26 &	33.80 &	32.96  \\
      &  &  &  &  -3.45 & \textbf{+4.24} &	\textbf{+3.56} &	\textbf{+1.10} &	\textbf{+1.44} &	\textbf{+1.63} &	-4.58 &	-2.07 &	-4.47 &	\textbf{+0.11} &	\textbf{+1.34} &	\textbf{+2.08} &	\textbf{+0.82} &	-1.04 &	\textbf{+0.05} \\
    \bottomrule
    \end{tabular}%
    }
    \end{table*}

    \begin{table*}[h]
    \centering
    \caption{{\textbf{Linear probing and K-NN classification} evaluation results of YFCC pretraining.}}
    \label{tab:YFCC_full_linear}
    \resizebox{\textwidth}{!}{%
    \begin{tabular}{cccccccccccccccccccccccc}
    \multirow{2}{*}{Arch.} & \multirow{2}{*}{Method} & \multirow{2}{*}{Data} & \multirow{2}{*}{{Epcoh (Rel.)}} &
    
    \multicolumn{2}{c}{CIFAR10} & \multicolumn{2}{c}{CIFAR100} & \multicolumn{2}{c}{DTD} & \multicolumn{2}{c}{Food101} & \multicolumn{2}{c}{OxfordIIITPet} & \multicolumn{2}{c}{RenderedSST2} & \multicolumn{2}{c}{StanfordCars} & \multicolumn{2}{c}{STL10} & \multicolumn{2}{c}{ImageNet} & \multicolumn{2}{c}{9 Dataset Avg.} \\
    
     &  &  &  &
    \multicolumn{1}{c}{Linear} & \multicolumn{1}{c}{KNN} & \multicolumn{1}{c}{Linear} & \multicolumn{1}{c}{KNN} & \multicolumn{1}{c}{Linear} & \multicolumn{1}{c}{KNN} & \multicolumn{1}{c}{Linear} & \multicolumn{1}{c}{KNN} & \multicolumn{1}{c}{Linear} & \multicolumn{1}{c}{KNN} & \multicolumn{1}{c}{Linear} & \multicolumn{1}{c}{KNN} & \multicolumn{1}{c}{Linear} & \multicolumn{1}{c}{KNN} & \multicolumn{1}{c}{Linear} & \multicolumn{1}{c}{KNN} & \multicolumn{1}{c}{Linear} & \multicolumn{1}{c}{KNN} & \multicolumn{1}{c}{Linear} & \multicolumn{1}{c}{KNN} \\

    \toprule
    \multirow{2}{*}{RN50} & CLIP & 15M & 32 (32.0) & 83.65 &	78.31 &	62.41 &	54.11 &	66.60 &	62.13 &	72.82 &	64.02 &	66.50 &	43.12 &	55.96 &	51.84 &	28.26 &	17.52 &	94.16 &	93.34 &	61.54 &	55.96 &	65.77 &	57.82  \\
     & ProtoCLIP & 14M & 8 (10.8) &  82.84 &	78.07 &	61.28 &	53.95 &	69.73 &	65.69 & 	69.62 &	60.13 &	67.10 &	45.52 &	54.48 &	51.46 &	27.67 &	19.67 &	94.16 &	93.28 &	62.13 &	56.72 &	65.45 &	58.28  \\
     &  &  &  &  -0.81 &	-0.24 &	-1.13 &	-0.16 &	\textbf{+3.14} &	\textbf{+3.56} &	-3.20 &	-3.89 &	\textbf{+0.60} &	\textbf{+2.40} &	-1.48 &	-0.38 &	-0.58 &	\textbf{+2.15} &	\textbf{0.00} &	-0.06 &	\textbf{+0.59} &	\textbf{+0.76} &	-0.32 &	\textbf{+0.46}  \\

     \midrule
    \multirow{2}{*}{RN101} & CLIP & 15M & 32 (32.0) & 83.37 &	78.89 &	62.07 &	53.42 &	67.50 &	62.02 &	74.08 &	65.85 &	66.18 &	46.25 &	56.95 &	52.33 &	27.70 &	19.14 &	94.81 &	95.03 &	63.10 &	57.85 &	66.19 &	58.98  \\
     & ProtoCLIP & 14M & 8 (10.8) &  83.34 &	79.57 &	60.58 &	55.76 &	70.43 &	66.28 & 	70.00 &	62.17 &	66.94 &	45.13 &	55.30 &	49.48 &	23.78 &	20.26 &	95.09 &	94.70 &	62.90 &	58.05 &	65.37 &	59.04 \\
     &  &  &  &  -0.03 & \textbf{+0.68} &	-1.49 &	\textbf{+2.34} &	\textbf{+2.93} &	\textbf{+4.26} &	-4.08 &	-3.68 &	\textbf{+0.76} &	-1.12 &	-1.65 &	-2.86 &	-3.92 &	\textbf{+1.12} &	\textbf{+0.27} &	-0.33 &	-0.20 &	\textbf{+0.19} &	-0.82 &	\textbf{+0.07} \\
    
    \bottomrule
    \end{tabular}%
    }
    
    \end{table*}

    \begin{table*}[h]
    \centering
    \caption{{\textbf{Image-text retrieval results} on MS-COCO dataset of YFCC pretraining.}}
    \label{tab:YFCC_full_retrieval}
    \resizebox{0.9\textwidth}{!}{%
    \begin{tabular}{ccccccccccc}
    Arch, & Method & Data & {Epcoh (Rel.)} &
    I2T R@1	& I2T R@5 &	I2T R@10 &	T2I R@1	& T2I R@5 &	T2I R@10 &	Mean Recall \\

    \toprule
    \multirow{2}{*}{RN50} & CLIP & 15M & 32 (32.0) & 26.46 &	52.12 &	63.82 &	16.47 &	37.23 & 49.11 &	40.87 \\
     & ProtoCLIP & 14M & 8 (10.8) &  30.20 &	55.08 &	66.54 &	16.89 &	37.93 &	49.40 &	42.67 \\
      &  &  &  &  \textbf{+3.74}	& \textbf{+2.96} & \textbf{+2.72} &	\textbf{+0.42} &	\textbf{+0.70} &	\textbf{+0.29} &	\textbf{+1.81} \\

     \midrule
    \multirow{2}{*}{RN101} & CLIP & 15M & 32 (32.0) & 29.26 &	55.12 &	67.16 &	17.69 &	39.15 &	50.57 &	43.16 \\
     & ProtoCLIP & 14M & 8 (10.8) &  31.46 &	57.78 &	69.28 &	18.17 &	39.87 &	51.46 &	44.67 \\
      &  &  &  &  \textbf{+2.20} & \textbf{+2.66} &	\textbf{+2.12} &	\textbf{+0.47} &	\textbf{+0.71} &	\textbf{+0.89} &	\textbf{+1.51} \\
    \bottomrule
    \end{tabular}%
    }
    \end{table*}

    \begin{table*}[h]
    \centering
    \caption{{\textbf{Object detection results} on MS-COCO dataset of YFCC pretraining.}}
    \label{tab:YFCC_full_COCO}
    \resizebox{0.8\textwidth}{!}{%
    \begin{tabular}{lc|cccccc|cccccc}
    Model & {Epcoh (Rel.)} & $AP^{b}$ & $AP^{b}_{50}$ & $AP^{b}_{75}$ & $AP^{b}_{S}$ & $AP^{b}_{M}$ & $AP^{b}_{L}$ & 
    $AP^{m}$ & $AP^{m}_{50}$ & $AP^{m}_{75}$ & $AP^{m}_{S}$ & $AP^{m}_{M}$ & $AP^{m}_{L}$  \\
    
    \toprule
    CLIP & 32 (32.0) &	36.5 &	58 &	39.4 &	21.9 &	39.4 &	47.1 &	34.2 &	55.1 &	36.3 &	16.7 &	36.7 &	49.4 \\
    ProtoCLIP & 8 (10.8) &	36.4 &	58.1 &	39.3 &	22.2 &	39.4 &	47.6 &	34.4 &	55.3 &	36.7 &	17.2 &	36.9 &	49.9 \\
    
    \bottomrule
    \end{tabular}%
    }
    \end{table*}

\subsection{Additional Experiments}\label{sec:Additional Experiments}
\subsubsection{Ablation on ProtoCLIP loss function}
    Here we study the effectiveness of each loss term in the ProtoCLIP loss function (Eq.~8). Table~\ref{tab:ablate_losses} summarizes the results of ImageNet linear probing accuracy. Adding $\mathcal{L}_{\text{Proto}}$ to $\mathcal{L}_{\text{CLIP}}$ improves representation grouping and improves linear accuracy by +3.78\%, introducing the external teacher further yields +1.79\% improvement.

\subsubsection{Ablation on ProtoCLIP Augmentation}
    Table~\ref{tab:ablate_augmentation} compares different data augmentation strategies. ``No Augmentation'' refers to using only the resize and crop with a random scale between 90\% and 100\%, which achieves the best image-text retrieval performance. Adding SimCLR augmentations degenerates all downstream performance. Our modified augmentations (``ProtoCLIP Augmentation'') improve the retrieval performance compared to ``SimCLR Augmentation'', and achieve the best ImageNet linear classification and zero-shot classification performance.


\begin{figure}[t]
\centering
\includegraphics[width=0.8\linewidth]{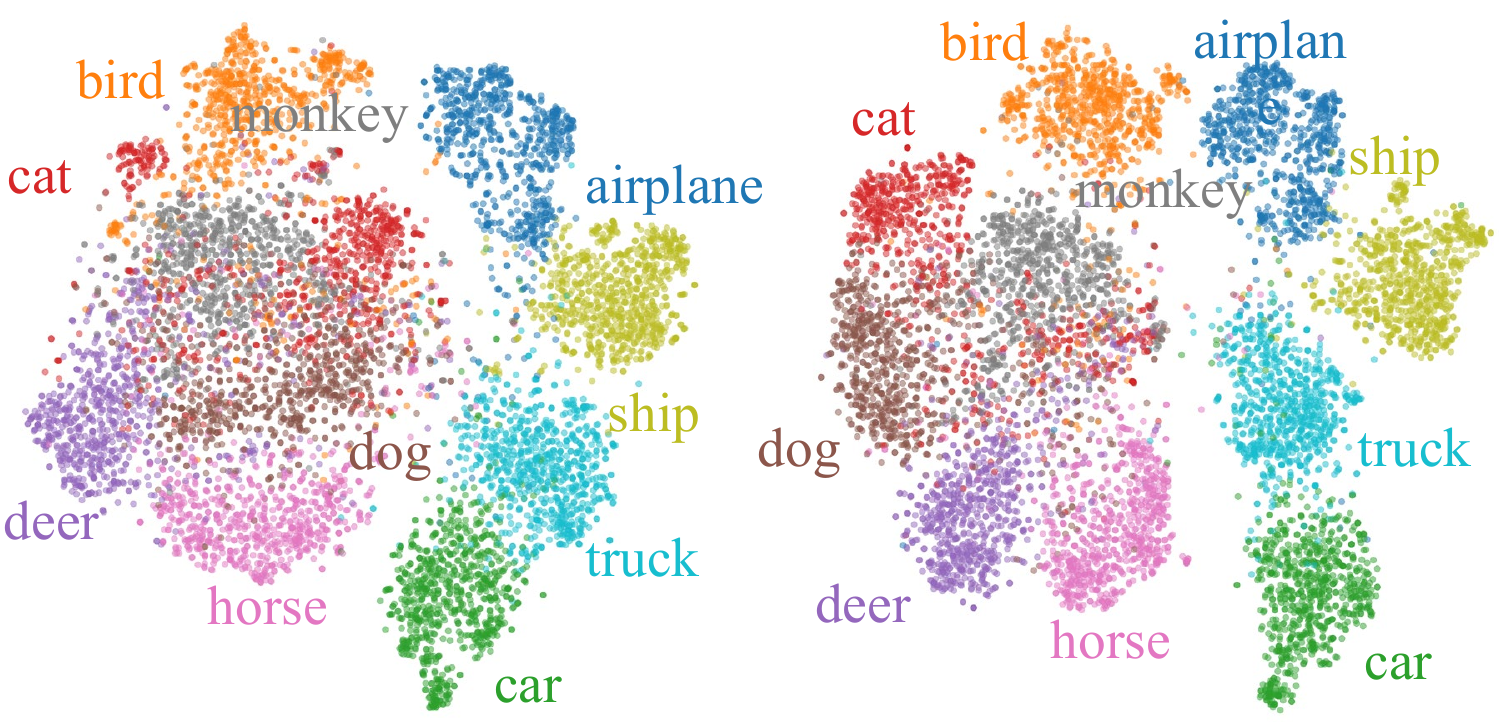}
        \caption{T-SNE visualizations of CLIP (left) and ProtoCLIP (right) representations on STL-10. ProtoCLIP yields a more clearly grouped representation space.}
        \label{fig:tsne}
\end{figure}

\subsubsection{Clustering Evaluation} 
\label{sec:Clustering Evaluation}
    Figure~\ref{fig:tsne} visualizes the learned representations of CLIP and ProtoCLIP via T-SNE~\cite{VanderMaaten2008Visualizing}. ProtoCLIP groups ``cat'', ``dog'', and ``monkey'' better. It also gives better separation between ``airplane'' and ``ship'', ``truck'' and ``car''. These observations can be proved by comparing clustering performance. We cluster the representations to 10 classes by \textit{K}-Menas and compare the obtained pseudo labels with ground truth labels. Representations of ProtoCLIP yields better adjusted rand index (0.673 $\to$ 0.732) and adjusted mutual information (0.744 $\to$ 0.788). 

    Next, we provide the clustering evaluation results of CLIP and ProtoCLIP trained on CC2.5M for 32 epochs. We extract test set image representations and perform \textit{K}-Means clustering to derive pseudo labels. The number of clusters (\textit{K}) is determined by the number of ground truth classes. We report the Adjusted Rand Index (ARI) and Adjusted Mutual Information (AMI) in Table~\ref{tab:clustering_evaluation}. ProtoCLIP outperforms CLIP in 8 out of 10 datasets.

    \begin{table}[t]
        \centering
        \caption{Ablation study of ProtoCLIP loss function}
        \label{tab:ablate_losses}
        \resizebox{0.8\linewidth}{!}{   
        \begin{tabular}{lc}
        \toprule
            Loss  Terms & ImageNet Linear Probing Accuracy\\
        \midrule
            $\mathcal{L}_{\text{CLIP}}$ & 40.98 \\
            $\mathcal{L}_{\text{Proto}}$ & 36.89 \\
            $\mathcal{L}_{\text{Proto}} + \mathcal{L}_{\text{CLIP}}$ & 44.76 \\
            \rowcolor[HTML]{DAE8FC} 
            $\mathcal{L}_{\text{Proto}} + \mathcal{L}_{\text{CLIP}} + \mathcal{L}^{\text{ external}}_{\text{Proto}}$  & \textbf{46.55} \\
        \bottomrule
        \end{tabular}}
        
    \end{table}

    \begin{table}[t]
        \centering
         \caption{\footnotesize Ablation study of data augmentation}
        \label{tab:ablate_augmentation}
        \resizebox{0.8\linewidth}{!}{    
        \begin{tabular}{cccc}
        \toprule
            \begin{tabular}[c]{@{}c@{}}Data \\ Augmentation \end{tabular} & \begin{tabular}[c]{@{}c@{}}ImageNet \\ Linear Acc.\end{tabular} & \begin{tabular}[c]{@{}c@{}}ImageNet \\ Zero-shot\end{tabular} & \begin{tabular}[c]{@{}c@{}}Mean \\ Recall \end{tabular} \\
        \midrule
            No Augmentation & 44.39 & 11.17 & \textbf{24.45} \\
            SimCLR Augmentation & 43.60 & 10.05 & 20.28 \\
            \rowcolor[HTML]{DAE8FC} 
            ProtoCLIP Augmentation & \textbf{46.55} & \textbf{11.96} & 21.65 \\
        \bottomrule
        \end{tabular}}
    \end{table}
    
    \begin{table*}
        \caption{Clustering evaluation results. ARI=Adjusted Rand Index, AMI=Adjusted Mutual Information}
        \label{tab:clustering_evaluation}
        \resizebox{\linewidth}{!}{%
        \begin{tabular}{cllllllllllllllllllllll}
        \toprule
         & \multicolumn{2}{c}{ImageNet} & \multicolumn{2}{c}{CIFAR   10} & \multicolumn{2}{c}{CIFAR100} & \multicolumn{2}{c}{STL10} & \multicolumn{2}{c}{Bidsnap} & \multicolumn{2}{c}{Country211} & \multicolumn{2}{c}{Flowers102} & \multicolumn{2}{c}{GTSRB} & \multicolumn{2}{c}{UCF101} & \multicolumn{2}{c}{Stanford   Cars} & \multicolumn{2}{c}{\textbf{10 Dataset Avg.}} \\
         & \multicolumn{1}{c}{ARI} & \multicolumn{1}{c}{AMI} & \multicolumn{1}{c}{ARI} & \multicolumn{1}{c}{AMI} & \multicolumn{1}{c}{ARI} & \multicolumn{1}{c}{AMI} & \multicolumn{1}{c}{ARI} & \multicolumn{1}{c}{AMI} & \multicolumn{1}{c}{ARI} & \multicolumn{1}{c}{AMI} & \multicolumn{1}{c}{ARI} & \multicolumn{1}{c}{AMI} & \multicolumn{1}{c}{ARI} & \multicolumn{1}{c}{AMI} & \multicolumn{1}{c}{ARI} & \multicolumn{1}{c}{AMI} & \multicolumn{1}{c}{ARI} & \multicolumn{1}{c}{AMI} & \multicolumn{1}{c}{ARI} & \multicolumn{1}{c}{AMI}& \multicolumn{1}{c}{ARI} & \multicolumn{1}{c}{AMI} \\
        \midrule
        CLIP 
        & 0.128 & 0.343 
        & \textbf{0.270} & \textbf{0.401} 
        & 0.130 & 0.340
        & 0.673 & 0.744 
        & 0.033 & 0.060 
        & 0.016 & 0.091 
        & 0.427 & 0.651 
        & \textbf{0.169} & \textbf{0.450} 
        & 0.305 & 0.579 
        & 0.020 & 0.103
        & 0.216 & 0.373 \\
        \rowcolor[HTML]{DAE8FC}
        ProtoCLIP 
        & \textbf{0.139} & \textbf{0.358} 
        & 0.263 & 0.393 
        & \textbf{0.138} & \textbf{0.365} 
        & \textbf{0.732} & \textbf{0.788} 
        & \textbf{0.042} & \textbf{0.073} 
        & \textbf{0.016} & \textbf{0.093} 
        & \textbf{0.479} & \textbf{0.688} 
        & 0.140 & 0.413 
        & \textbf{0.360} & \textbf{0.619} 
        & \textbf{0.021} & \textbf{0.107} 
        & \textbf{0.233} & \textbf{0.390}\\
        \bottomrule
        \end{tabular}
        }
        
    \end{table*}

\subsubsection{Efficiency Analysis of Episodic Training.} 
    We analyze the time consumption of each step in the episodic training. On a 8$\times$2080Ti machine with 60 CPUs and 300G RAM, one episode takes an average of 6 minutes. As shown in Figure~\ref{fig:profiling}, episodic training of ProtoCLIP requires an additional feature extraction step compared to CLIP, which takes 31.0\% time. \textit{K}-Means clustering takes only 2.3\% of time, since the number of samples in an episode is not too large. Smaller episodes also save the total \textit{K}-Means time cost since its time complexity grows superlinearly $O(m^{d_h \times K + 1} )$ along the number of samples $m$ to be clustered. 
    The PBT step takes negligible time.

\begin{figure}[t]
\centering
    \includegraphics[width=0.55\linewidth]{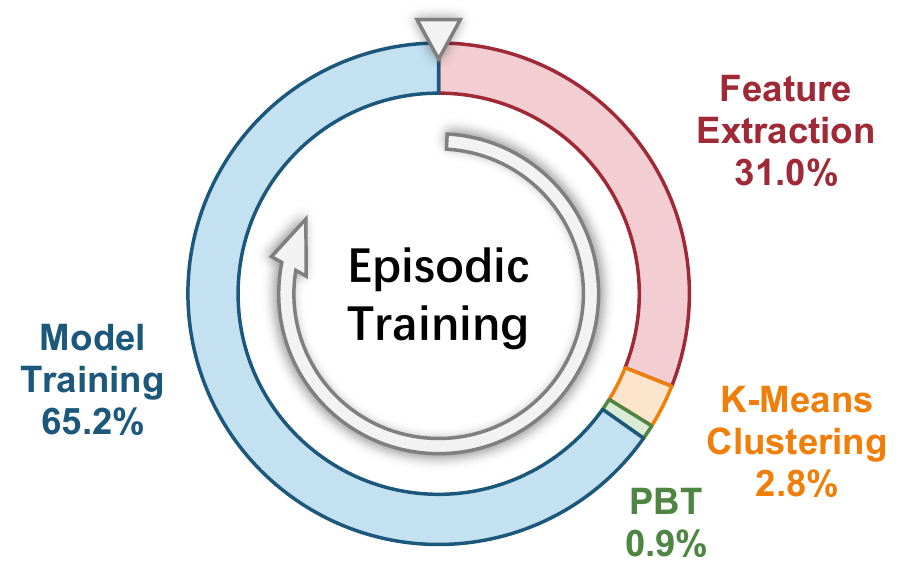}
    \caption{{Time profiling of the episodic training strategy. Although ProtoCLIP takes an additional 34.8\% time cost for each epoch, the overall normalized efficiency is improved by the prototypes constructed during the feature extraction, clustering, and PBT steps.}}
    \label{fig:profiling}
\end{figure}

\section{Conclusion}
    We have shown that in addition to representation alignment, representation grouping is also an important characteristic of contrastive language image pretraining. The InfoNCE objective groups representations together via randomly emerged anchors, which we found unstable and sensitive to the modality gap. We set up stable and efficient representation grouping via prototypical discrimination (ProtoCLIP) and alleviated the modality gap issue by PBT. PBT also enabled us to introduce an external teacher for additional supervision. Empirical results proved that all these novel designs bring improvements to downstream performance.

\normalem
\bibliographystyle{ieeetr}
\bibliography{ProtoCLIP.bib}
\vfill

 




\begin{IEEEbiography}[{\includegraphics[width=1in,height=1.25in,clip,keepaspectratio]{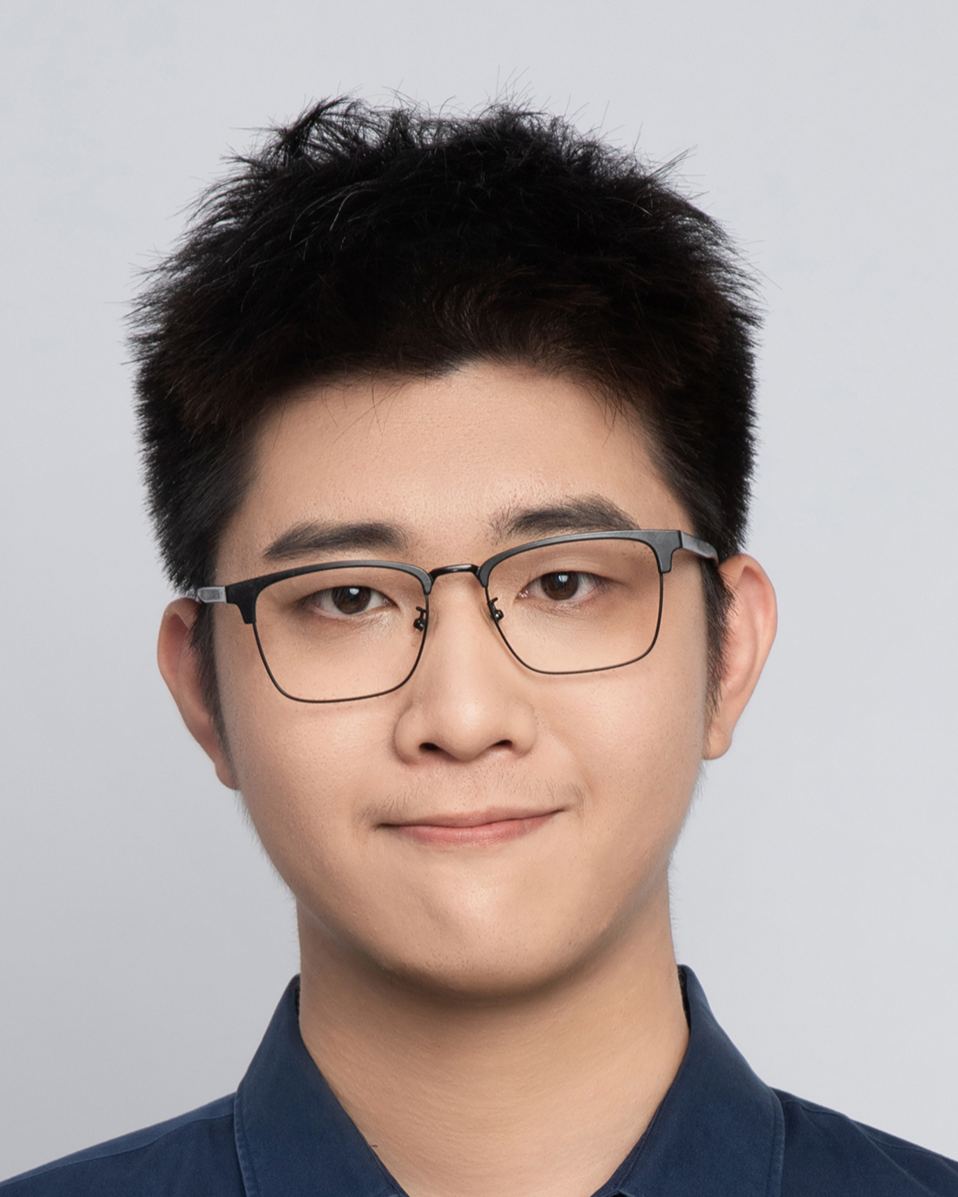}}]{Delong Chen} received his BEng degree in computer science from Hohai University, Nanjing, China in 2021. He interned at Megvii research and Xiaobing.AI during 2021-2023, and is currently a Ph.D. student at Centre for Artificial Intelligence Research (CAiRE) of Hong Kong University of Science and Technology (HKUST). His research includes vision-language learning, multi-modality, and self-supervised learning. He received the Best Demo award in IEEE ICME'21, Best Paper award in AAAI'23 Inaugural Summer Symposium, and the LTDL Best Dataset paper in IJCAI'21.
\end{IEEEbiography}

\begin{IEEEbiography}[{\includegraphics[width=1in,height=1.25in,clip,keepaspectratio]{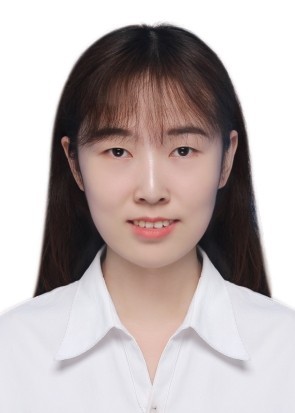}}]{Zhao Wu} received the BS, MS degrees from Xidian University (XDU), Xi'an, China, in 2017 and 2020, respectively. She is now at Megvii Research. Her research interests include computer vision and machine learning, in particular, the area of general object detection, knowledge distillation, multi-modal learning.
\end{IEEEbiography}

\begin{IEEEbiography}[{\includegraphics[width=1in,height=1.25in,clip,keepaspectratio]{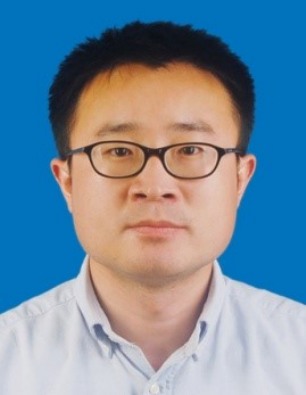}}]{Fan Liu}
 is currently a professor at Hohai University. He received his B.S. degree and Ph.D. degree from Nanjing University of Science and Technology (NUST) in 2009 and 2015.  From September 2008 to December 2008, he studied at Ajou University in South Korea. From February 2014 to May 2014, he worked at Microsoft Research Asia. His research interests include computer vision, pattern recognition, and machine learning. Dr. Liu serves as a reviewer of \emph{IEEE TNNLS},  \emph{IEEE TKDE},  \emph{ACM TIST},  \emph{Information Sciences},  \emph{Neurocomputing},  \emph{Pattern Analysis and Application} and an executive director of Jiangsu association of Artificial Intelligence (JSAI).
\end{IEEEbiography}

\begin{IEEEbiography}[{\includegraphics[width=1in,height=1.25in,clip,keepaspectratio]{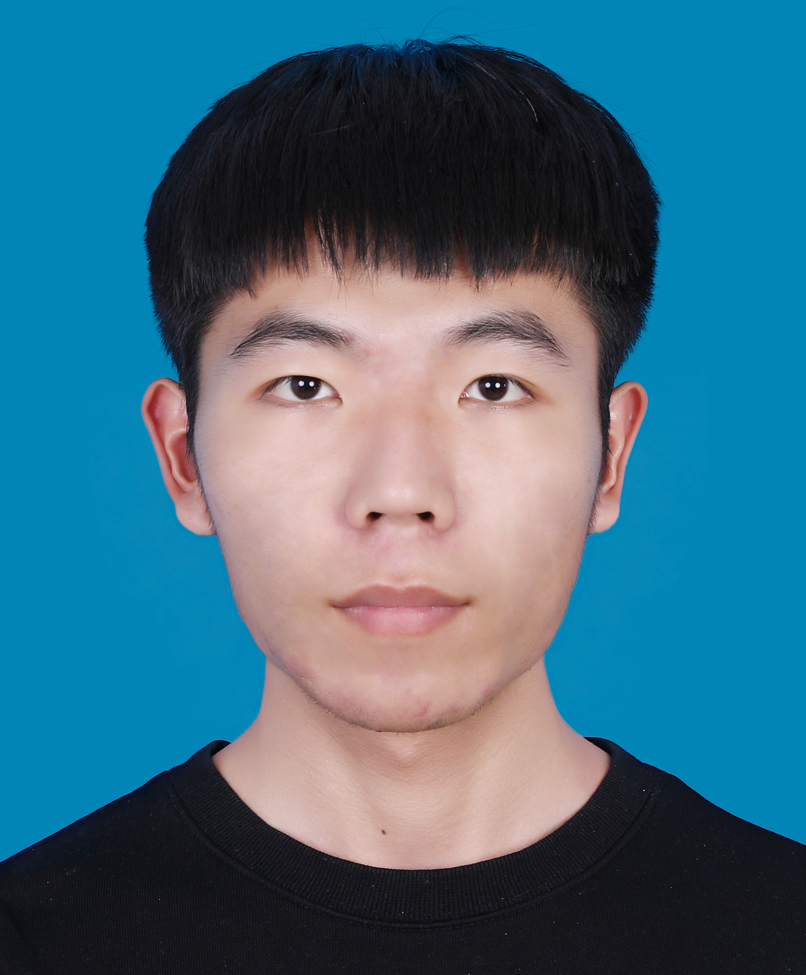}}]{Zaiquan Yang} is currently pursing the MS degree in computer science in Beihang University. He received his BS degree from the Beijing Institute of Technology in 2020. He is interested in computer vision, weakly supervised learning and cross-modal learning. 
\end{IEEEbiography}

\begin{IEEEbiography}[{\includegraphics[width=1in,height=1.25in,clip,keepaspectratio]{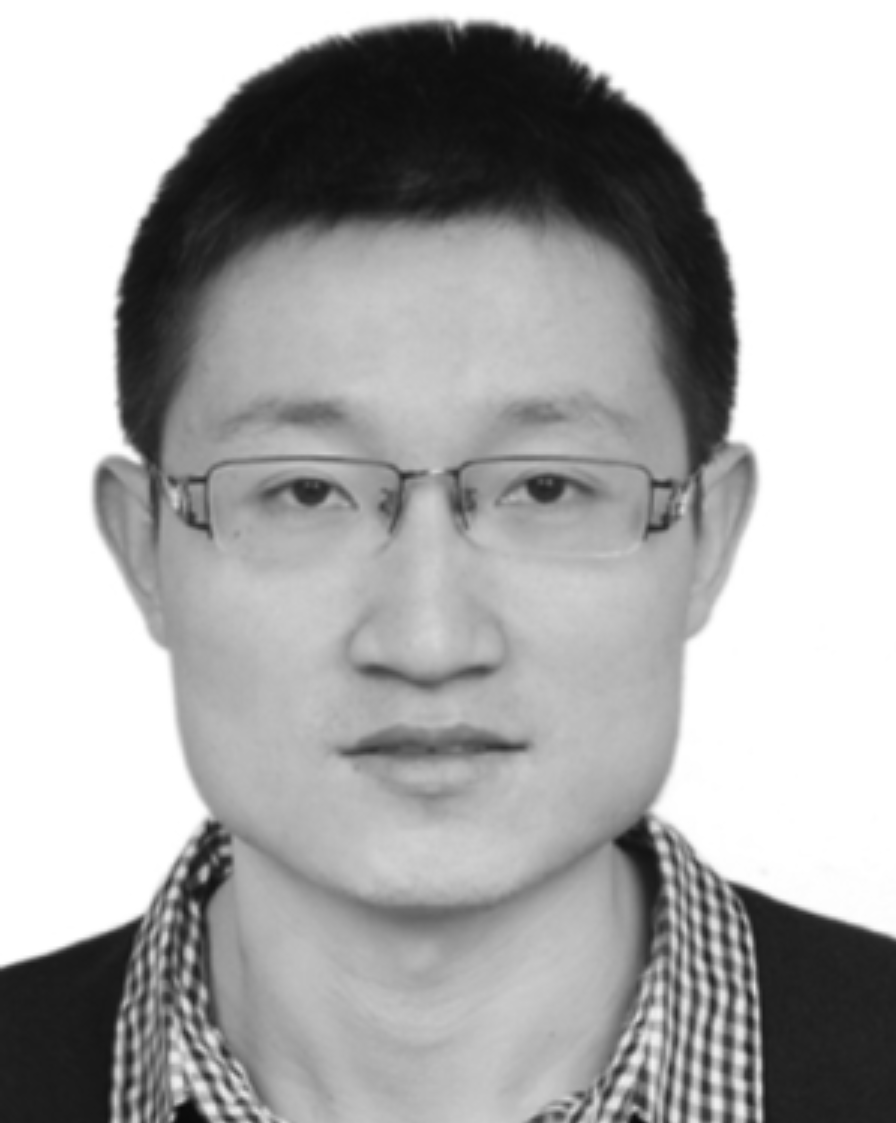}}]{Shaoqiu Zheng}received the BEng degree from the Department of Underwater Acoustic Engineering, Northwestern Polytechnical University, in June 2010, and the PhD degree from the Department of Computer Science and Technology, Peking University, in June 2015. 

He is currently a senior research engineer at the Nanjing Research Institute of Electronic Engineering. His research interests include intelligent command and control system, deep reinforcement learning, swarm intelligence.
\end{IEEEbiography}

\begin{IEEEbiography}[{\includegraphics[width=1in,height=1.25in,clip,keepaspectratio]{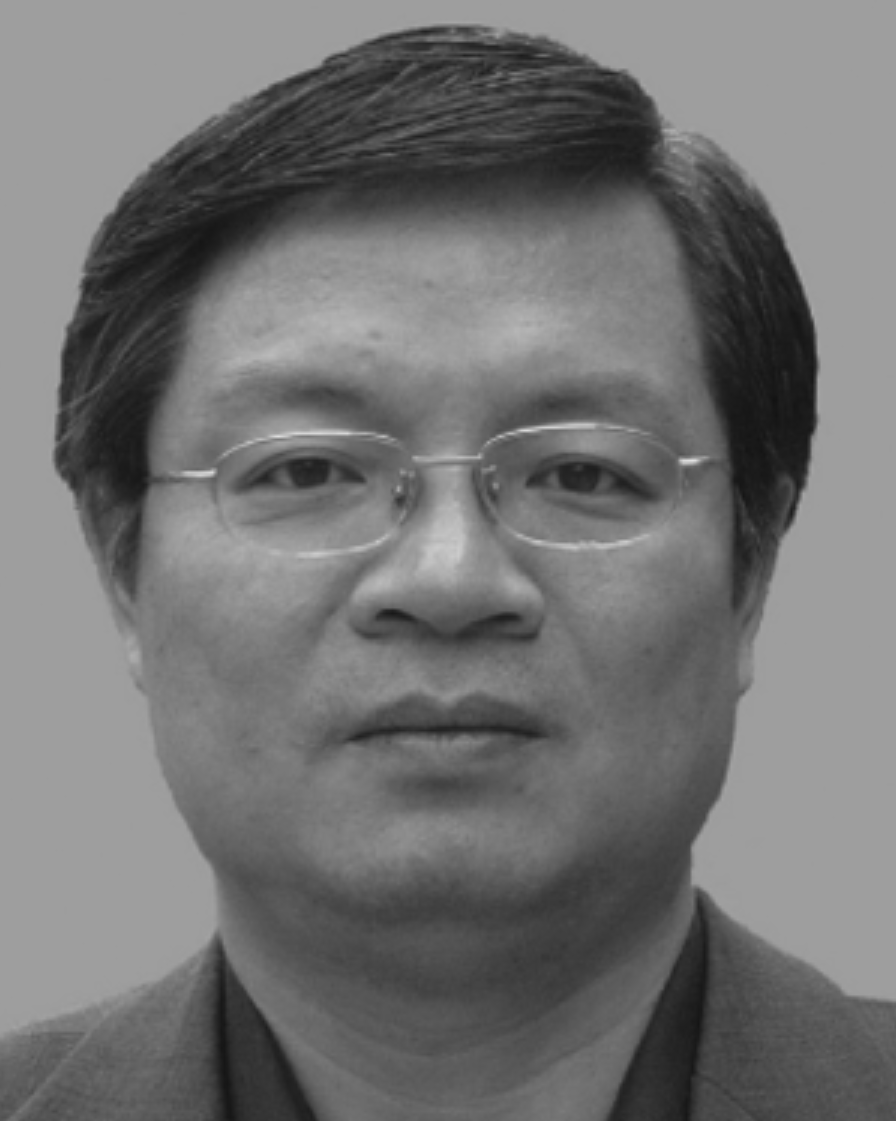}}]{Ying Tan} (SM’02) received the B.Eng., M.S., and Ph.D. degrees from Southeast University, Nanjing, China, in 1985, 1988, and 1997, respectively.
He is a full professor and Ph.D. advisor with the School of Electronics Engineering and Computer Science, Peking University, Beijing, China, and also the Director of Computational Intelligence Laboratory. He is the inventor of the fireworks algorithm. His current research interests include computational intelligence, swarm intelligence, data mining, pattern recognition, and intelligent information processing for information security. He has published over 260 papers in refereed journals and conferences in the above areas, authored/co-authored ten books and 12 book chapters, and received three invention patents.

Prof. Tan was a recipient of the Second-Class Natural Science Award of China in 2009. He serves as an Editor-in-Chief of the International Journal of Computational Intelligence and Pattern Recognition, and as an Associate Editor of the IEEE Transactions on Cybernetics, the IEEE Transactions on Neural Networks and Learning Systems (TNNLS), the International Journal of Swarm Intelligence Research (IJSIR), and the International Journal of Artificial Intelligence (IJAI). He also served as an Editor for Springer’s Lecture Notes on Computer Science for over 20 volumes, and a Guest Editor for several refereed journals, including the IEEE/ACM Transactions on Computational Biology and Bioinformatics, Information Science, Softcomputing, Neurocomputing, IJSIR, IJAI, Bioinspiration \& Biomimetics, and Computer Journal. He is the Founder General Chair of the ICSI International Conference series. He has been a member of the Emergent Technologies Technical Committee of the IEEE Computational Intelligence Society since 2010.
\end{IEEEbiography}

\begin{IEEEbiography}[{\includegraphics[width=1in,height=1.25in,clip,keepaspectratio]{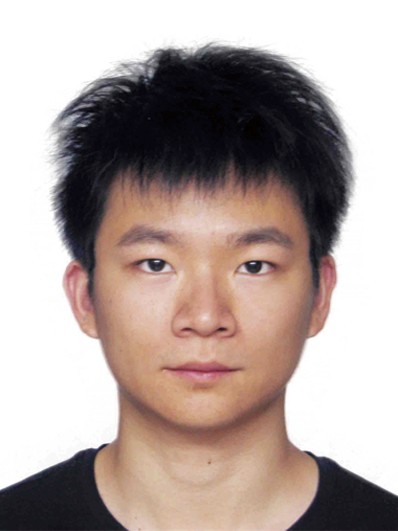}}]{Erjin Zhou} received the BS degree in Electronic Engineering from the Tsinghua University, in 2015. Immediately following, he joined Megvii Research, and has been working in the fields of computer vision and machine learning. His primary research interests are deep learning based face recognition, landmark localization, general object detection, and building real-world working computer vision systems in the cloud/mobile based solutions.
\end{IEEEbiography}

\end{document}